\documentclass[preprint,3p]{elsarticle}

\usepackage{amssymb}

\usepackage{makecell}
\usepackage{graphicx}
\usepackage{svg}
\usepackage{xcolor}
\usepackage{booktabs}
\usepackage{multicol}
\usepackage{tabularx}
\usepackage{threeparttable}
\usepackage{booktabs}
\usepackage{longtable}
\usepackage{hyperref}
\usepackage{amsthm, amsmath}
\usepackage{algorithm}
\usepackage{algpseudocode}
\usepackage{placeins}
\usepackage{amsfonts}
\usepackage{pgfplots}
\usepackage{subcaption}
\usepackage{todonotes}

\usepackage{mlearning}

\newcommand{\beq}{\begin{equation}}
\newcommand{\eeq}{\end{equation}}

\begin{document}

\begin{frontmatter}

\title{Flow-based conditional cardiac anatomy generation for virtual cohorts}
\author[inst1,inst2]{Konstantinos Kevopoulos}
\author[inst1,inst2,inst3]{Beatrice Moscoloni}
\author[inst1]{Benjamin Alheit}
\author[inst2,inst4]{Cameron Beeche}
\author[inst2]{Julio A. Chirinos}
\author[inst5]{Alexander Heinlein}
\author[inst1]{Mathias Peirlinck\corref{cor}}
\ead{mplab-me@tudelft.nl}
\cortext[cor]{Correspondence:}

\affiliation[inst1]{organization={Dept. BioMechanical Engineering, Delft University of Technology, Delft, The Netherlands}}
\affiliation[inst2]{organization={Division of Cardiovascular Medicine, Hospital of the University of Pennsylvania, Philadelphia, PA, USA}}
\affiliation[inst3]{organization={BioMMeda – Institute for Biomedical Engineering and Technology, Ghent University, Ghent, Belgium}}
\affiliation[inst4]{organization={Dept. Bioengineering, University of Pennsylvania, Philadelphia, PA, USA}}
\affiliation[inst5]{organization={Delft Institute of Applied Mathematics, Delft University of Technology, Delft, The Netherlands}}

\begin{abstract}
Cardiac digital twin research is moving from subject-specific anatomical replicas toward virtual cohorts that represent clinically relevant population subgroups. 
Yet access to representative imaging-derived anatomy datasets remains limited by cohort size, subgroup sparsity, and data-sharing constraints. 
Conditional generative models could help address this gap, but virtual cohorts are useful only if they preserve realistic, metadata-dependent anatomical variability. 
Existing cardiac anatomy generators largely rely on conditional variational autoencoders (cVAEs), which couple representation learning and metadata conditioning through a shared regularized latent prior. 
We introduce CAN-FLOW, a two-step Conditional ANatomy generation framework based on normalizing FLOWs that first learns geometry-only latent representations of diffeomorphic cardiac shape momenta and then models their sex-, age-, and body-mass-index-dependent distribution with a conditional normalizing flow.
We trained CAN-FLOW on 2,208 healthy UK Biobank subjects and compared it with cVAEs across regularization strengths. 
CAN-FLOW generated plausible stochastic biventricular anatomies that better reproduced clinical phenotype distributions, metadata-dependent trends, subgroup variability, point-cloud coverage, and high-dimensional shape variability. 
Together, these results establish CAN-FLOW as a shareable framework for generating realistic, stochastically varying, metadata-conditioned biventricular anatomies for virtual cohort construction and in silico clinical trial workflows.
\end{abstract}


\end{frontmatter}

\section{Introduction}
\label{sec:introduction}

Cardiac digital twin research has traditionally focused on subject-specific anatomical models for personalized simulations of electrophysiology, hemodynamics, mechanics, and device--heart interactions \cite{Jiang2016,Augustin2016,CorralAcero2020,Peirlinck2021sexdiff, Peirlinck2021precmed,Zingaro2024,Martinez2026,Tikenogullari2023,Salvador2024EPtwinCHD, qian2025developing}.
Increasingly, however, digital twin and in silico clinical trial efforts are moving beyond single-subject models toward virtual cohorts or virtual populations that represent the anatomical variability of broader target populations \cite{Niederer2020}.
Such cohorts are needed to evaluate model predictions, therapies, and medical devices across the range of anatomies in which they are intended to operate \cite{Aycock2024, Pathmanathan2024}.
For these applications, generating plausible average anatomies is insufficient: virtual cohorts must preserve clinically relevant anatomical phenotype distributions and the morphological variability that governs how interventions and devices affect varying populations.

One route towards creating such cohorts is to build more anatomical models directly from images.
Cardiac image-to-model pipelines have advanced rapidly, and recent open-access frameworks now make it possible to process cardiac magnetic resonance images into biventricular meshes with increasing efficiency, reproducibility, and practical deployability \cite{Dillon2025BivMe, Ugurlu2025, Doste2026, GonzalezMartin2024,Gsell2026, govil2023deep}. 
This progress lowers the technical barrier for cardiac digital twin development. 
It does not, however, remove the underlying data-availability barrier. 
Many research groups face restricted access to large imaging resources. 
Major population imaging cohorts are usually governed by project-specific application and approval procedures. 
Patient-specific images, segmentations, and derived anatomical models also remain constrained by privacy, consent, data ownership, and data-use constraints, which limit broad distribution \cite{Peloquin2020,littlejohns2020uk}. 
As a result, there is a mismatch between the growing need for population-level virtual cohorts and the limited availability of representative cardiac anatomy collections that can be readily shared, reused, and interrogated across institutions.

Synthetic anatomical cohort generation offers a complementary solution to this problem. 
Instead of exchanging individual image-derived geometries, a trained generative model could be used to generate synthetic anatomies that resemble the population from which it was learned \cite{Kong2024, Sorensen2024, kadry2024probing}. 
For cardiac digital twin applications, however, this is only useful if the generated anatomies preserve the relevant anatomical features of the real population. 
The generative framework must reproduce inter-subject variability, subgroup-specific trends, and the rare but plausible anatomies found in the tails of the anatomical distribution of the real population.
It must also do so conditionally. 
In other words, a useful generator should not merely produce realistic biventricular shapes. 
It should also learn how the distribution of cardiac anatomy changes with clinically interpretable subject-level characteristics such as sex, age, and body size \cite{qiao2023cheart,Qiao2025}.

This conditional view is also relevant beyond cohort generation. 
If a model can learn the expected distribution of biventricular anatomies for a given sex, age, and body size, it can then provide a richer reference against which an individual heart can be compared. 
Current clinical and population imaging studies often summarize cardiac anatomy using scalar phenotypes such as ventricular volumes, myocardial mass, or wall thickness. 
These measures are essential because they are interpretable and clinically standardized. 
Yet they compress anatomy into a small set of numbers and therefore cannot fully capture regional morphology or high-dimensional biventricular shape variation \cite{bruse2016statistical,beetz2022interpretable}. 
Conditional anatomical models offer a complementary perspective: they can characterize how an individual anatomy deviates from the expected shape distribution of a matched reference population, potentially yielding shape-based biomarkers that go beyond conventional scalar phenotypes.

Existing generative models of cardiac anatomy have largely relied on variational autoencoder (VAE) frameworks \cite{kingma2013auto, sohn2015learning,beetz2022interpretable,beetz2022multi,Kuznetsov2021,peng2023generating}. 
VAEs compress each high-dimensional anatomy into a lower-dimensional latent representation and reconstruct it through a decoder. 
To generate new anatomies without an input anatomy, the model must define a distribution from which latent representations can be sampled. 
VAEs therefore regularize the encoded representations toward a known prior, typically a standard Gaussian, so that samples from this prior are compatible with the latent codes used to train the decoder. 
Conditional VAEs (cVAEs) extend this formulation by providing subject characteristics, such as sex, age, or body size, to both the encoder and decoder, allowing generated anatomies to depend on these characteristics \cite{beetz2021generating, dou2023conditional, qiao2023cheart, Qiao2025}. 
In the standard cVAE formulation, however, the prior remains shared across anatomies and does not vary with subject characteristics. 
This shared regularization facilitates prior-based sampling but it also constrains how differently the latent distributions of distinct metadata groups can vary. 
When this regularization is strong, cVAEs may retain less anatomy-specific information and underrepresent subgroup-specific anatomical variability, particularly in the tails of the distribution.
For virtual cohort generation, these tails matter because less frequent but physiologically plausible anatomies may be important when testing whether an intervention or device design succeeds across anatomical configurations.

To address this limitation, we introduce CAN-FLOW, a framework for Conditional ANatomy generation with normalizing FLOWs. 
Rather than jointly learning a regularized latent representation and its dependence on subject characteristics, CAN-FLOW separates representation learning from conditional distribution modeling. 
First, each anatomy is represented by a lower-dimensional set of momenta using large-deformation diffeomorphic metric mapping. 
Rather than representing the anatomy directly, these momenta describe the deformation that maps a template anatomy to the target anatomy, providing a geometrically aware representation with anatomical correspondence. 
An autoencoder then further compresses the momenta into a lower-dimensional, geometry-only representation. 
Second, a conditional normalizing flow learns how the distribution of these anatomical representations varies with sex, age, and body mass index. 
By separating these two learning tasks, CAN-FLOW models the metadata-dependent distribution of anatomical representations directly, without imposing a shared, conditioning-independent prior during representation learning \cite{xiao2019generative, stein2022probabilistic, kaechele2025comparison}.

In this study, we apply CAN-FLOW to generate synthetic end-diastolic biventricular cardiac anatomies conditioned on sex, age, and body mass index, using a healthy subset of the UK Biobank cardiac imaging cohort. 
We compare CAN-FLOW with cVAE baselines trained on the same data.
More specifically, we assess whether the generated cohorts satisfy five critical requirements for virtual cohort generation: anatomical plausibility, clinical phenotype fidelity, preservation of sex-, age-, and body-mass-index-dependent phenotype distributions, subgroup-specific anatomical variability, and high-dimensional shape variability beyond scalar phenotypic measures.
Together, these analyses test whether CAN-FLOW and cVAE-based conditional generation can produce realistic biventricular virtual cohorts for population-specific cardiac digital twin development and in silico trial design.

\section{Results}
\label{sec:results}

\begin{figure}[h!]
\centering
  \includegraphics[width=\linewidth]{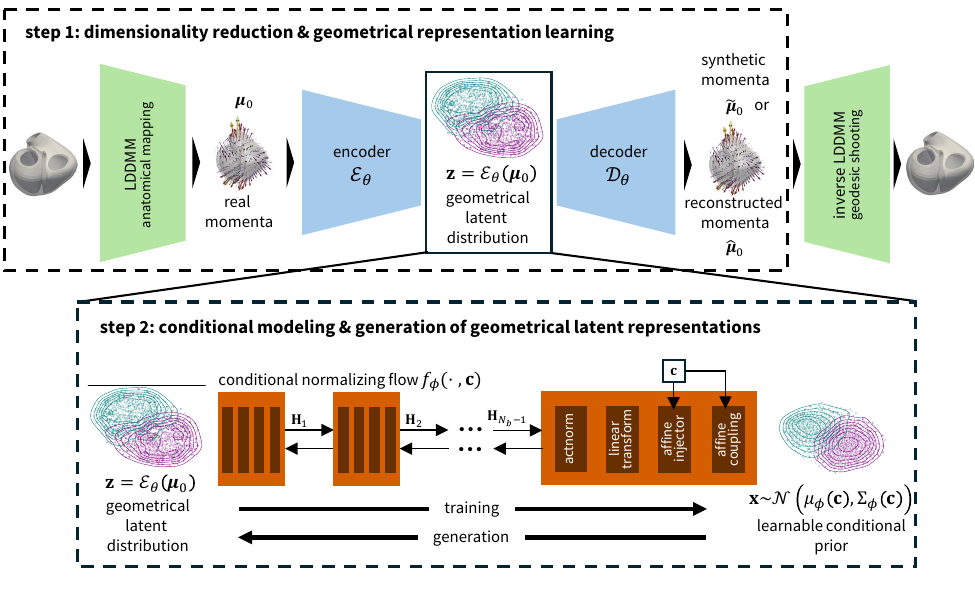}
  \caption{\textbf{CAN-FLOW generates metadata-conditioned biventricular anatomies through a two-step representation and distribution learning strategy.}
  Subject-specific end-diastolic biventricular surface meshes are first represented as diffeomorphic momenta $\boldsymbol{\mu}_0$ through LDDMM, and then embedded into lower-dimensional geometry-only representations $\mathbf{z}=\mathcal{E}_{\theta}(\boldsymbol{\mu}_0)$ space using an autoencoder with learnable weights $\{\mathcal{E}_{\theta}, \mathcal{D}_{\theta}\}$.
  A conditional normalizing flow $f_{\phi}$ then learns the metadata-dependent distribution of these latent anatomical representations.
  During generation, the flow samples from a learnable conditional prior according to sex, age, and body mass index (metadata $\mathbf{c}$), and generates synthetic latent representations through the inverse flow mapping. Then, the decoder maps those to synthetic momenta $\tilde{\boldsymbol{\mu}}_0$; and geodesic shooting maps the momenta back to synthetic biventricular surface anatomies.}
  \label{fig:fig_1}
\end{figure}

We developed CAN-FLOW to generate virtual cohorts of end-diastolic biventricular anatomies conditioned on demographic (sex, age) and anthropometric (body mass index - BMI) characteristics, from here on out referred to as \textit{metadata}.
Figure~\ref{fig:fig_1} showcases CAN-FLOW's two-step strategy: a metadata-agnostic autoencoder first learns geometry-only latent representations, after which a conditional normalizing flow learns the metadata-dependent distribution of these latent representations.
Once trained, our normalizing-flow model is employed to sample from the learned latent distribution, and synthetic latent representations are transformed into anatomies through a decoding process. 
Our input anatomies were standardized biventricular surface meshes derived from cardiac magnetic resonance images and represented through diffeomorphic momenta, following the protocol described by Moscoloni et al. \cite{moscoloni2025unveiling}.

We evaluated CAN-FLOW's ability to produce synthetic cohorts that preserve the anatomical variability observed in the real population.
All models were trained on the same UK Biobank-derived cohort disclosed in~\ref{subsec:dataset}, and conditioned on sex, age, and body mass index.
We compared CAN-FLOW with cVAE baselines using complementary analyses.
First, we assessed whether the generated anatomies were visually plausible and covered the overall shape distribution of the real cohort.
Second, we tested whether synthetic cohorts were able to reproduce clinically interpretable anatomical phenotypes, including 
left ventricular end-diastolic volume (LVEDV), 
right ventricular end-diastolic volume (RVEDV), 
myocardial mass, 
RVEDV-to-LVEDV ratio, 
long-axis length of the LV, and 
sphericity of the LV.
Third, we evaluated whether the generated anatomies preserved phenotype trends across sex, age, and BMI.
Finally, we quantified whether generated shape cohorts retained more detailed geometric anatomical variability across varying population subgroups.

Detailed clinical and higher-dimensional shape variability evaluation metrics
were introduced in Methods \ref{sec:methods}, and their definitions were provided in~\ref{subsec:Appendix_performance_metrics}.
To avoid selecting a weak baseline for the state-of-the-art comparison, we trained cVAE models over a wide range of values for $\beta$. 
This hyperparameter weighted the latent-space regularization term in the training objective and was introduced here and described in Section \ref{sec:methods}. 
We reported the two strongest performing variants in the main text.
These variants corresponded to $\beta = 10^{-2}$ and $\beta = 10^{-3}$. These values provided the most competitive trade-off between reconstruction fidelity and generative diversity.
Results for the remaining $\beta$ values are reported in \ref{sec:AppendixSupplementaryResults}.

\subsection{CAN-FLOW generates plausible biventricular anatomies}
\begin{figure}[!htbp] 
\centering
  \includegraphics[width=\linewidth]{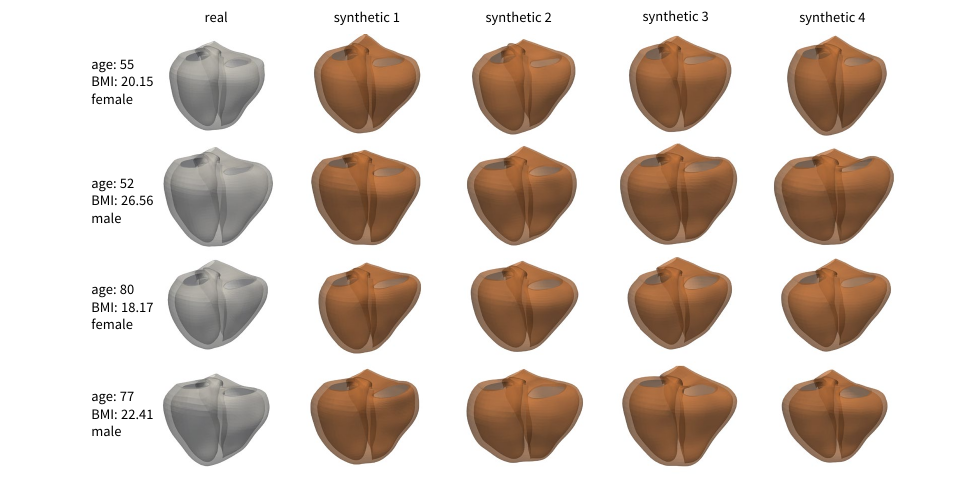}
  \caption{\textbf{Real and synthetic biventricular anatomies, according to subject-specific metadata}. 
  For subjects with different metadata, the real anatomies and four synthetic ones, generated by CAN-FLOW, are shown. 
  The real anatomy is illustrated in gray color, and the synthetic ones in dark orange.
  }
  \label{fig:fig_2}
\end{figure}

We first evaluated whether CAN-FLOW generates visually plausible biventricular anatomies. 
For each of four test subjects with different metadata, we generated a synthetic anatomy conditioned on the same metadata as the corresponding real subject. 
These examples should be interpreted as metadata-conditioned samples, not as subject-specific reconstructions. 
Differences of the generated anatomies from the displayed real anatomies are therefore expected and reflect the sampled anatomical variability. 
As illustrated in Figure~\ref{fig:fig_2}, the generated anatomies reproduced key structural characteristics of the real cohort, including comparable overall size and similar basal LV shape. 
They also retained visible subject-to-subject variation in chamber length and width. 
Consistent with reported sex-related differences in cardiac anatomy \cite{moscoloni2025unveiling, StPierre2022sexmatters}, male-conditioned samples tended to be larger than female-conditioned samples.

\clearpage

\subsection{CAN-FLOW preserves anatomical variability within targeted subgroups}
\begin{figure}[h!]
\centering
  \includegraphics[width=\linewidth]{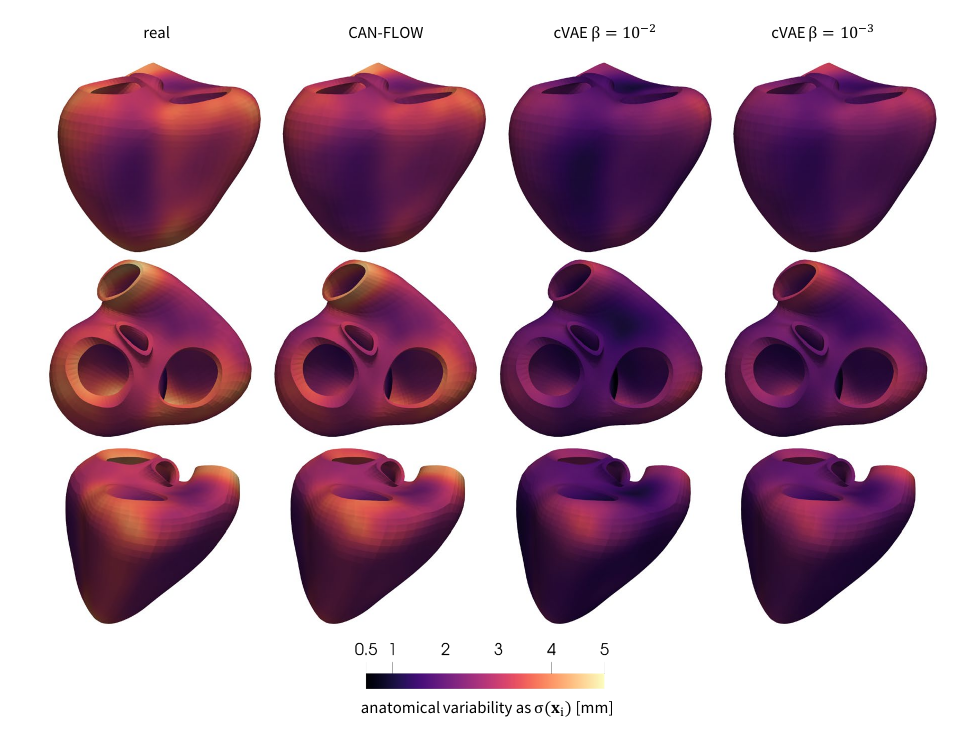}
  \caption{\textbf{Node-wise standard deviation of real and synthetic cohorts, for a specific population subgroup}.
  The population subgroup under consideration includes males over 58 years of age, and the real cohort consists of 674 subjects.
  Node-wise variability is visualized on the LDDMM template anatomy, with $\mathbf{x}_i$ denoting the position of each surface-mesh point.}
\label{fig:fig_3}
\end{figure}

A central use case for conditional anatomy generation is the construction of targeted virtual cohorts with specific demographic or anthropometric characteristics. 
We therefore evaluated whether the models preserved anatomical variability within an illustrative target subgroup, defined as males older than 58 years. 
Using our CAN-FLOW and the traditional cVAE frameworks respectively, we generated subgroup-specific synthetic cohorts. 
We then quantified local anatomical variability as the node-wise standard deviation of corresponding surface-point positions across the biventricular meshes. 
Larger values indicate regions where anatomies differ more strongly across subjects. 
We visualize this distance-based variability measure in Figure~\ref{fig:fig_3}. 
The cVAE baselines produced uniformly lower spatial variability than the real subgroup. 
This indicates that they generated overly homogeneous cohorts and failed to capture part of the anatomical variability observed among real subjects. 
In contrast, CAN-FLOW reproduced spatial patterns of variation that more closely matched the real cohort, including variability in the outflow regions, ventricular apex, and ventricular surfaces.

\subsection{Synthetic cohorts preserve clinically relevant phenotype distributions across the entire cohort}
\begin{figure}[h!]
\centering
  \includegraphics[width=\linewidth]{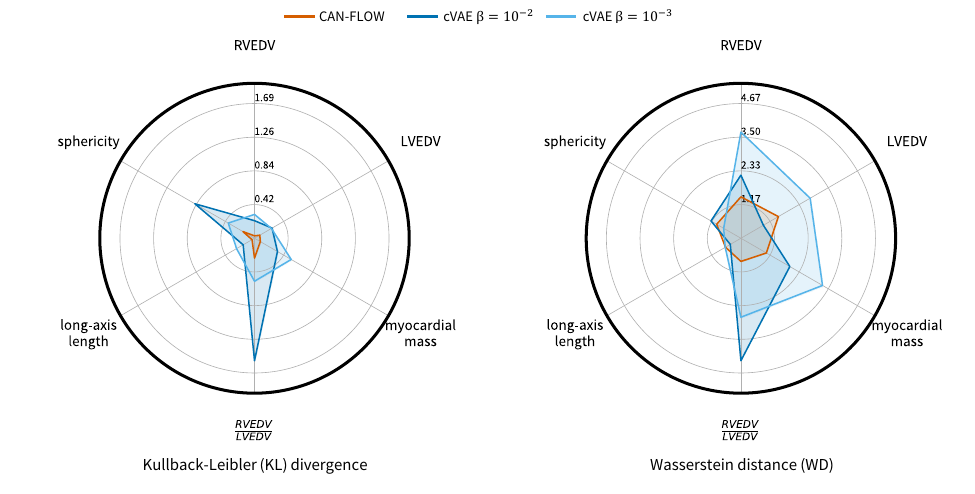}
  \caption{\textbf{Comparison of real and synthetic clinical phenotype distributions using KL divergence and Wasserstein distance}.
  Radar plots show KL divergence and Wasserstein distance between real and synthetic distributions for each clinical phenotype and generative model. Smaller values indicate better agreement between the generated and real phenotype distributions.}
  \label{fig:fig_4}
\end{figure}

To assess whether generated anatomies preserved clinically interpretable mesh-derived phenotypes, we compared real and synthetic distributions of LVEDV, RVEDV, myocardial mass, RVEDV-to-LVEDV ratio, LV long-axis length, and LV sphericity using Wasserstein distance \cite{arjovsky2017wasserstein} and Kullback-Leibler (KL) divergence \cite{cover1999elements}.
For both metrics, lower values indicate closer agreement between the generated and real phenotype distributions.
As shown in Figure~\ref{fig:fig_4}, CAN-FLOW achieved the lowest KL divergence for all clinical phenotypes. 
It also substantially outperformed the cVAE baselines in Wasserstein distance for RVEDV, myocardial mass, and RVEDV-to-LVEDV ratio, while remaining competitive for the remaining phenotypes.

\begin{figure}[h!]
\centering
  \includegraphics[width=\linewidth]{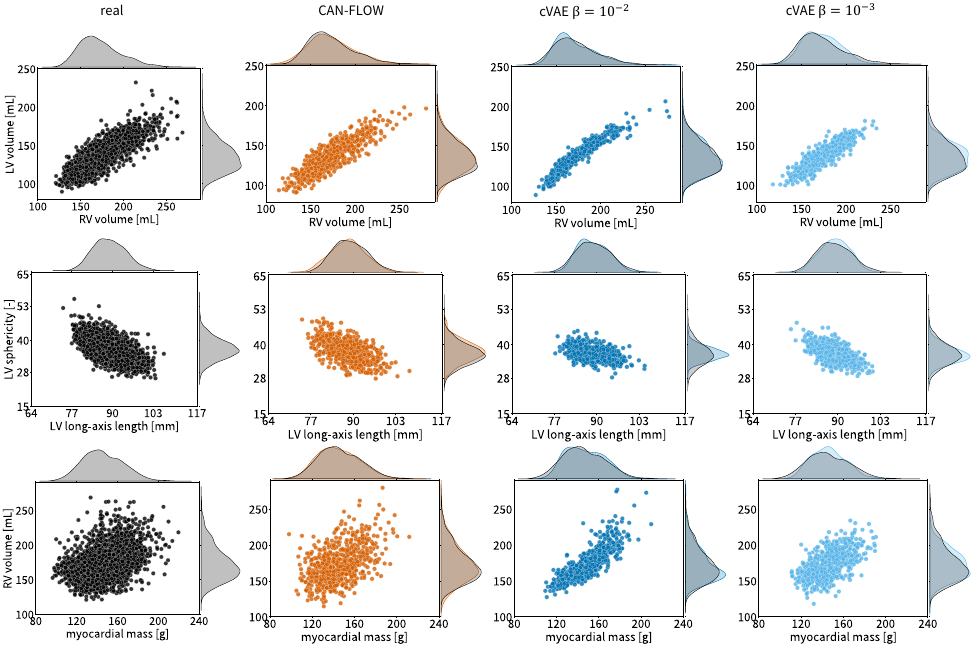}
  \caption{\textbf{Joint and marginal distributions of clinical phenotypes across the whole cohort}. 
  Each row shows the joint distribution of one phenotype pair for real anatomies, CAN-FLOW-generated anatomies, and cVAE-generated anatomies. 
  Scatter plots show the occupied clinical phenotype space, while marginal distributions are estimated using kernel density estimation. For each generative model, the corresponding real total cohort and synthetic marginal distributions are overlaid for comparison.}
  \label{fig:fig_5}
\end{figure}

Joint phenotype distributions of LV-to-RV volumes, LV sphericity to LV long-axis length, and RV volume to myocardial mass respectively, as shown in Figure~\ref{fig:fig_5}, supported the same conclusion. 
CAN-FLOW more faithfully reproduced the real phenotypic space, including the tails of the distributions and less frequent but physiologically plausible phenotype combinations. 
In contrast, cVAE-generated anatomies were more tightly clustered around the anatomical means, leaving parts of the real phenotype variability space underrepresented.
This suggests that CAN-FLOW better preserves less frequent but physiologically plausible combinations of the illustrated clinical phenotypes, which are important for virtual cohort construction.

\subsection{CAN-FLOW recovers metadata-dependent clinical phenotypes}
\begin{figure}[h!]
\centering
  \includegraphics[width=\linewidth]{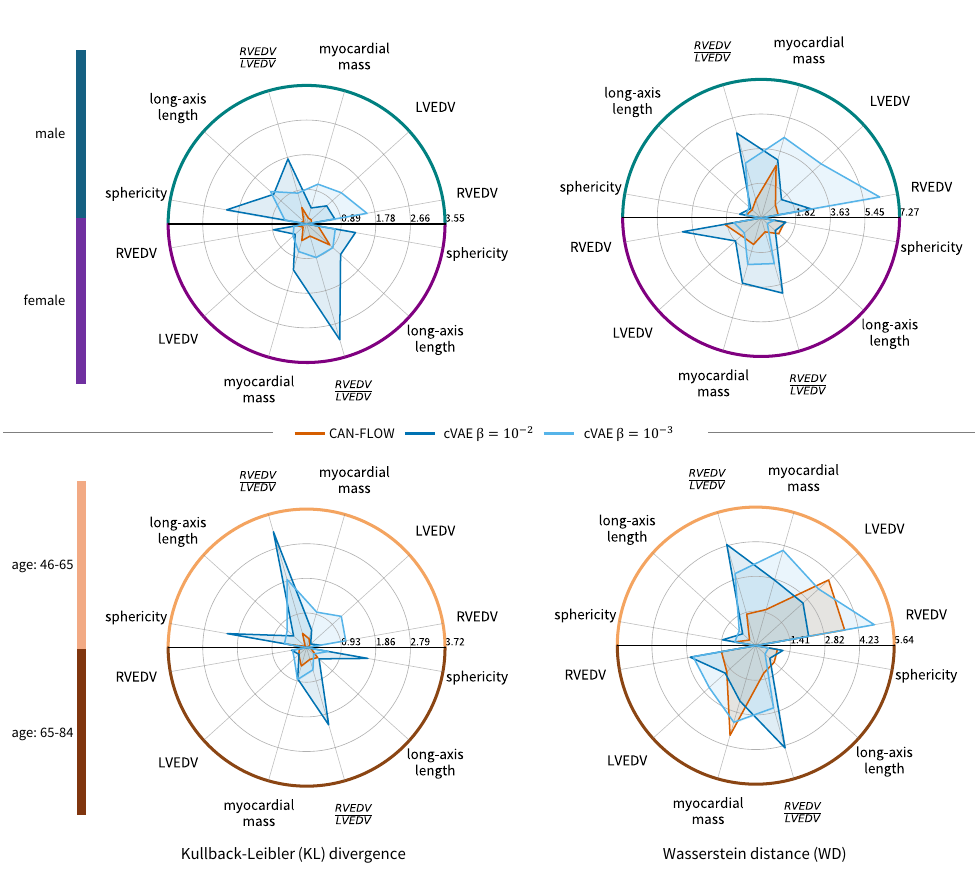}
  \caption{\textbf{KL divergence and Wasserstein distance for sex- and age-conditioned clinical phenotype distributions.} 
  Radar plots compare real and synthetic phenotype distributions within metadata-defined subgroups. 
  Smaller values, plotted closer to the center, indicate closer agreement between the generated and real subgroup-stratified distributions.}
  \label{fig:fig_6}
\end{figure}

The preceding analysis assessed whether generated anatomies reproduced the overall clinical phenotype distributions. 
For conditional virtual cohort generation, however, matching the aggregate cohort is not sufficient. 
A useful conditional anatomy generator should also preserve shape phenotype distributions across metadata-defined subgroups. 
We therefore compared real and synthetic phenotype distributions after stratifying the cohorts by sex and age. 
Figure~\ref{fig:fig_6} shows that, across these conditional subgroup distributions, CAN-FLOW achieved the lowest KL divergence in all cases and the best or competitive Wasserstein distances compared with the strongest cVAE baselines.

\begin{figure}[h!]
\centering
  \includegraphics[width=\linewidth]{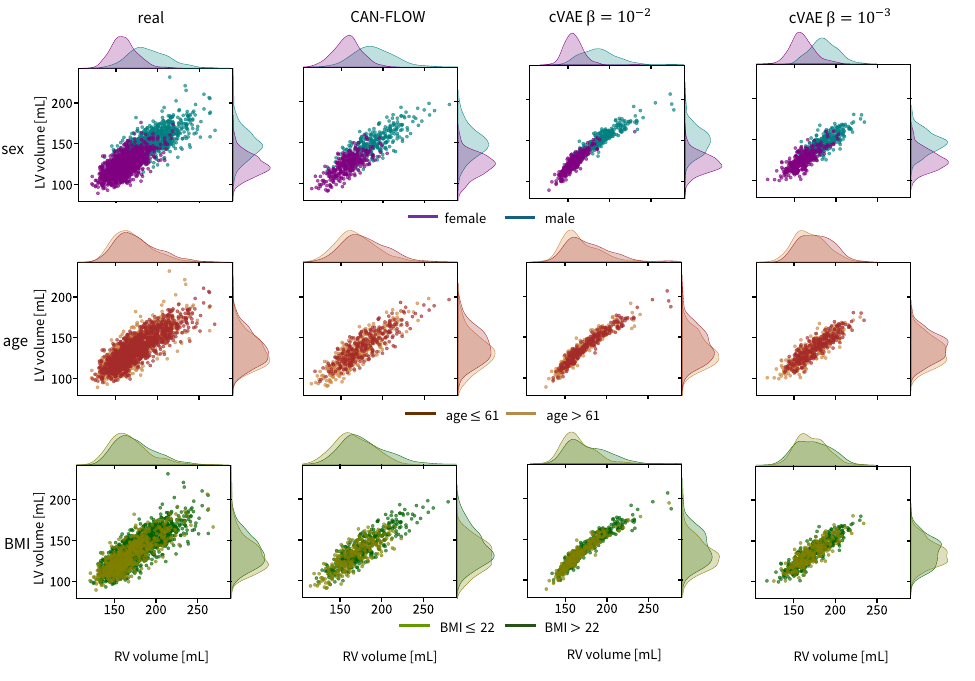}
  \caption{\textbf{Subgroup-stratified LVEDV-RVEDV distributions in real and synthetic cohorts.}
  Scatter plots compare the joint LVEDV-RVEDV distributions of real anatomies and synthetic anatomies generated by CAN-FLOW and the cVAE baselines. 
  Cohorts are stratified by sex, age group, and BMI group, with each metadata category divided into two groups and colored accordingly. 
  Marginal distributions are estimated using kernel density estimation.}
  \label{fig:fig_7}
\end{figure}

To further inspect whether the models preserved metadata-dependent phenotype trends, we visualized the joint LVEDV--RVEDV distributions after stratifying real and synthetic cohorts by sex, age, and BMI (Figure~\ref{fig:fig_7}). 
Age and BMI were each partitioned into two groups, with thresholds selected to retain sufficient sample sizes in both groups and to produce visually distinct phenotype distributions. 
CAN-FLOW most closely reproduced the real cohort across these strata. The improvement was clearest for sex-conditioned distributions, where CAN-FLOW better captured both the modes and tails of the LVEDV and RVEDV distributions. 
Age- and BMI-stratified distributions showed stronger overlap in the real data, making visual interpretation less direct; nevertheless, CAN-FLOW maintained closer agreement with the real marginal distributions than the cVAE baselines.

\subsection{CAN-FLOW generated cohorts retain shape variability beyond scalar phenotype metrics}
Anatomies with similar clinical phenotypes can still differ in detailed geometric features, including regional wall thickness, outflow morphology, and septal curvature. 
We therefore asked whether the generated cohorts preserved anatomical variability beyond the scalar phenotype metrics analyzed above. 
To do so, we quantified variability directly in the diffeomorphic shape momenta representation space, see Section~\ref{sec:CMR-to-mesh}, using principal component analysis (PCA).

For each PCA coefficient, we compared two distances. 
First, we computed the Wasserstein distance between two equally sized subsets of the real cohort, which provides a reference level of natural inter-cohort shape variability. 
Second, we replaced one real subset with a generated cohort from CAN-FLOW or a cVAE baseline and computed the corresponding real-to-synthetic Wasserstein distance. 
The ratio between these two Wasserstein distances was used as the comparative metric, as detailed in \ref{subsubsec:AppendixPCAEXPMETHODS}. 
Ratios close to one indicate that the generated cohort reproduces the variability expected between two real cohort subsets, whereas ratios farther from one indicate poorer agreement with real inter-cohort variability.

We considered two settings. 
In the \textit{metadata-agnostic} setting, we tested whether each model captured overall shape variability within the real cohort. 
In the \textit{sex-specific} setting, we used the Wasserstein distance between real male and real female PCA coefficient distributions as the reference sex-specific variability. 
For the CAN-FLOW and cVAE models, we then replaced the real female cohort with a generated female cohort and computed the Wasserstein distance between the real male and generated female PCA coefficient distributions. 
The plotted ratio compares these two distances. 
Values close to one indicate that the generated female cohort reproduces the real female distributional separation from the male cohort.
In both the metadata-agnostic and sex-specific settings, we analyzed the leading 65 PCA coefficients, which together cover more than 95\% of the real dataset variability.

\begin{figure}[h!]
\centering
  \includegraphics[width=\linewidth]{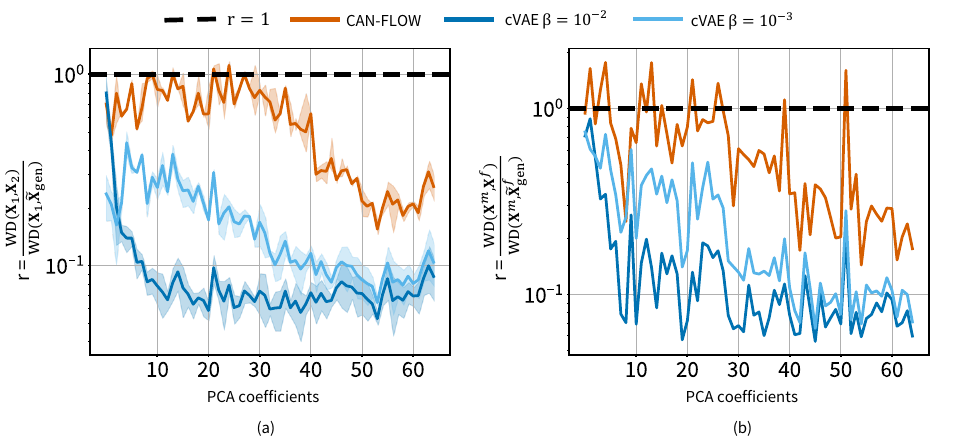}
  \caption{\textbf{PCA-based comparison of anatomical variability in shape momenta space.} 
  Anatomical variability is summarized by the ratio $r$ between real-to-real and real-to-synthetic Wasserstein distances computed from PCA shape mode coefficient distributions. 
  Values close to one indicate that the generated cohort reproduces the variability expected between real cohort subsets. 
  (a) \textit{Metadata-agnostic} setting: real inter-cohort variability is estimated from two random real subsets, and compared with variability between real and generated cohorts. Confidence intervals show $\pm \sigma$ over five random splits. 
  (b) \textit{Sex-specific} setting: the reference distance is computed between real male and real female PCA coefficient distributions. For each model, this reference is compared with the distance between real male and generated female PCA coefficient distributions.
  In both settings, the leading 65 PCA coefficients are shown, together explaining more than 95\% of the variance in the real dataset.}
  \label{fig:fig_8}
\end{figure}

In Figure~\ref{fig:fig_8}, CAN-FLOW reproduced the real variability profile more closely than the cVAE baselines in both settings. 
Its inter-cohort Wasserstein distance ratios approached one for a larger number of leading PCA shape momenta coefficients, indicating closer agreement between CAN-FLOW-generated shape modes and real anatomical variability. 
In contrast, the cVAE baselines showed consistently lower ratios for most leading coefficients, consistent with compressed shape variability. 
Additional support is provided in Figure~\ref{fig:fig_appendix_4}, where the marginal distributions of the ten most important PCA coefficients are compared for the real, CAN-FLOW-, and cVAE $(\beta=10^{-2})$-based cohorts. 
CAN-FLOW more closely matched the real PCA coefficient distributions, particularly in the tails, suggesting better preservation of less frequent anatomical variations. 
These findings indicate that CAN-FLOW preserves not only global clinical phenotype distributions, but also geometrically detailed modes of anatomical variation required for realistic biventricular virtual cohorts.

\subsection{CAN-FLOW better spans the real anatomical distribution in biventricular point-cloud geometry}
Having evaluated anatomical variability in shape momenta space, we next asked whether the same pattern was reflected directly in surface point-cloud space cohort coverage.
We assessed real-to-synthetic agreement using two complementary point-cloud metrics: minimum matching distance (MMD) and Coverage (Cov) \cite{yang2019pointflow}.
MMD measures nearest-neighbor geometric fidelity, with lower values indicating that real anatomies can be closely matched by generated anatomies. Cov measures distributional breadth, with higher values indicating that the synthetic cohort spans a larger fraction of the real anatomical distribution.

\begin{figure}[h!]
\centering
  \includegraphics[width=\linewidth]{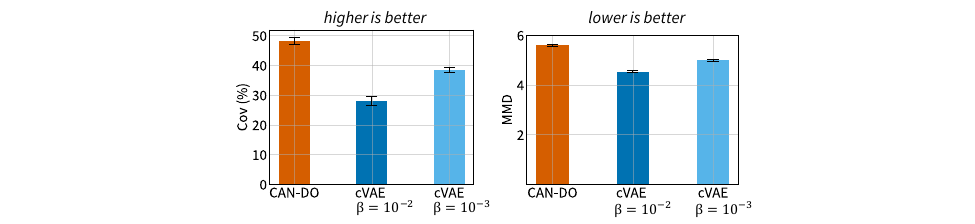}
  \caption{\textbf{Point-cloud comparison of real and synthetic cohort coverage and geometric fidelity.} 
  Synthetic cohorts $S_g$ contain $|S_g|=600$ anatomies, balanced by sex with 300 female and 300 male samples. 
  For comparison, 600 anatomies are subsampled from the real cohort $S_r$ so that $|S_r|=|S_g|$. 
  Bars show the mean and standard deviation across four independent subsampling repetitions. 
  Higher coverage indicates broader representation of the real anatomical distribution, whereas lower MMD indicates closer nearest-neighbor geometric agreement with real anatomies.}
  \label{fig:fig_9}
\end{figure}

Figure~\ref{fig:fig_9} shows that CAN-FLOW achieved approximately 10\% higher coverage than the cVAE baselines, indicating that it captured a broader range of real anatomical variability. 
In contrast, the cVAE baselines achieved lower MMD values, indicating closer nearest-neighbor similarity to individual real anatomies. 
This trade-off is consistent with the reconstruction-focused behavior of the cVAE baselines, which can favor anatomies close to frequently represented regions of the training distribution, whereas CAN-FLOW better reproduced the distributional spread of the cohort.

\section{Discussion}
\label{sec:discussion}
We introduced CAN-FLOW to address a central requirement for synthetic generation of organ anatomies: 
synthetic cohorts should contain plausible anatomies while preserving the subpopulation-specific, i.e. metadata-conditioned, distributional breadth of true subject-specific anatomy variability.
In the present cardiac application, these subpopulations were stratified by sex, age, and body mass index.
CAN-FLOW separates geometry-only representation learning from metadata-conditioned density estimation. 
An autoencoder first learns compact representations of diffeomorphic momenta which capture the essential variations in cardiac shape, and a conditional normalizing flow then models metadata-dependent distribution of these representations. 
Across clinical phenotype-based, diffeomorphic shape momenta-based, and high-dimensional surface point-cloud evaluations, CAN-FLOW generated virtual cohorts that more faithfully preserved the variability of a real UK Biobank-derived cohort than traditional cVAE baselines.

The most important improvement of CAN-FLOW is not simply the generation of plausible biventricular shapes, but the preservation of distributional spread. 
The cVAE baselines achieved lower MMD values, indicating close nearest-neighbor similarity to individual real anatomies. 
However, they consistently showed lower coverage and reduced anatomical variability. 
In contrast, CAN-FLOW better reproduced the tails of the phenotypic distributions, achieved higher cohort shape phenotype and shape distribution coverage, and more closely matched PCA-based shape and node-wise distance variability patterns. 
For virtual cohort applications, this trade-off is important: a cohort of highly average anatomies may appear realistic while still underrepresenting the anatomical variability required for representative in silico clinical studies.

Beyond matching aggregate cohort distributions, CAN-FLOW also better preserved conditioning-dependent anatomical distributions. 
Synthetic cohorts generated by CAN-FLOW more closely reproduced phenotype trends with respect to sex, age, and body mass index.
Additionally, they better captured anatomical variability within a targeted subgroup of the overall population. 
This distinction is important because conditional virtual cohorts should reflect the expected anatomy of the subgroup under study, not only the global population average. 
Preserving subgroup-specific anatomical heterogeneity is particularly relevant when virtual cohorts are used to test robustness across plausible anatomical configurations.

The improved performance of CAN-FLOW can be interpreted in terms of its two-step formulation. 
In cVAEs, the encoder and decoder are trained jointly while the latent distribution is regularized toward a shared, typically Gaussian, prior. 
This formulation supports sampling, but may underrepresent metadata-dependent anatomical variability.
CAN-FLOW instead leaves the geometry-only latent space unconstrained during autoencoder training and learns its conditional distribution separately with a normalizing flow. 
We showed this separation helps to preserve anatomical variability while still enabling controllable generation. 
Supplementary ablation studies further support this interpretation, showing that CAN-FLOW retained its advantage across different flow hyperparameters and latent-space dimensionalities, as reported in~\ref{sec:appendixAblationStudies} and~\ref{sec:appendixAblationStudiesLatentDim}. 
These analyses suggest that the observed gains arise primarily from the two-step generative formulation rather than from a highly optimized flow architecture.

A further application of conditional anatomy generation is the construction of normative shape references. 
Because CAN-FLOW learns a distribution of healthy biventricular anatomies conditioned on subject metadata, it could potentially be used to quantify how far an observed anatomy deviates from the expected healthy shape distribution for a given subject profile. 
For example, the anatomy of a new subject could be compared against a metadata-matched synthetic healthy cohort to derive a conditional shape-distance score. 
Such scores may eventually support the detection or stratification of abnormal cardiac remodeling. 
The present study does not evaluate diagnostic performance, and this application will require validation in independent diseased cohorts with appropriate clinical labels and outcome data.

Several aspects of our study define the scope within which these results should be interpreted. 
First, CAN-FLOW was trained and evaluated on healthy UK Biobank participants, and therefore learns a normative distribution of end-diastolic biventricular anatomy within this source population rather than a universal reference for healthy cardiac anatomy. 
As with any population imaging cohort, the learned distribution may reflect the demographic composition, recruitment patterns, imaging protocol, and health profile of the underlying cohort. Second, the present implementation considers static end-diastolic anatomy only. 
This scope matches the central aim of our study: to learn conditional population-level variability in cardiac shape, rather than cardiac motion.
Extending the framework to time-resolved cardiac anatomy would be a distinct generative task, requiring not only plausible anatomy at each phase but also phase-consistent motion, temporal smoothness, and physiologically coherent deformation trajectories \cite{Moscoloni2025fimhlddmmstrain}.
Third, all generated anatomies inherit the assumptions of the upstream cardiac magnetic resonance imaging to mesh and diffeomorphic registration pipeline, including potential segmentation, mesh-fitting, and correspondence errors.

Future work should evaluate CAN-FLOW as part of broader trustworthy synthetic-data pipelines for virtual human organ twin research. 
This will require validation in external cohorts, diseased populations, and underrepresented subgroups to assess whether conditional synthetic anatomies preserve clinically relevant variability beyond the healthy UK Biobank population studied here \cite{Pentenga2023shapemorphing, Salvador2024EPtwinCHD, Tikenogullari2023}.
By generating metadata-conditioned anatomies without directly sharing individual image-derived meshes, this approach may provide a pathway toward more accessible and privacy-conscious virtual anatomy cohorts. The same strategy may be extended to other anatomical structures or multi-organ cardiac models, provided that suitable shape representations and preprocessing pipelines are available. More broadly, these results support conditional anatomical shape generation as a step toward trustworthy, population-specific virtual cohorts for cardiac digital twins and in silico trials.

\clearpage
\section{Methods}
\label{sec:methods}
We first describe the UK Biobank cohort and image-to-shape preprocessing, then define the CAN-FLOW generative framework, the cVAE comparison baselines, and the evaluation metrics used to compare real and synthetic anatomies.

\subsection{UK Biobank cohort and anatomical data}
\label{subsec:dataset}
This study was conducted using imaging, anthropometric, and demographic data provided by the UK Biobank \cite{littlejohns2020uk}. 
From a dataset comprising more than 100,000 participants, we selected healthy individuals with no reported history of cardiovascular disease, hypertension, respiratory disease, diabetes mellitus,
hyperlipidemia, hematologic disease, renal disease, rheumatologic disease, malignancy, symptoms of chest pain or dyspnea \cite{petersen2016uk}.
We further excluded individuals who were current or former smokers, as well as those with a BMI $\geq$ 30 $kg/m^2$.
The resulting healthy subset included 1222 females and 1052 males.
We processed the end-diastolic cardiac magnetic resonance images using the automated segmentation and mesh-fitting pipeline described in \cite{Dillon2025BivMe}. 
Visual inspection of the resulting meshes revealed a small number of unrealistic cases, likely caused by segmentation artifacts. 
After removing these outliers, the final dataset consisted of 2208 individuals; 1200 females and 1008 males. 
Further details regarding outlier removal are provided in~\ref{subsec:Appendix_OutlierRemoval}.
We divide this dataset into three non-overlapping subsets for training, validating, and testing the conditional shape-generation model: $70\%$ for training, $15\%$ for validation, and $15\%$ for testing.

\subsection{Image-to-mesh preprocessing and diffeomorphic shape representation}\label{sec:CMR-to-mesh}
We converted the end-diastolic cardiac magnetic resonance images into geometrically consistent 3D shape representations before training the generative model. 
For each individual, we used the fully automated segmentation and mesh-fitting pipeline of Dillon et al. \cite{Dillon2025BivMe}. 
This pipeline classifies cardiac magnetic resonance imaging views, segments the cardiac chambers and myocardium using a pretrained 3D nnU-Net, and fits a mesh to the resulting contours through iterative diffeomorphic registration \cite{mauger2018iterative}. 
We then aligned the derived meshes with the iterative closest point algorithm \cite{zhang1994iterative, schroeder1998visualization}, following a procedure similar to \cite{moscoloni2025unveiling}, to remove pose differences before anatomical correspondence was estimated.

After extracting and aligning the 3D biventricular surface meshes, we applied large deformation diffeomorphic metric mapping (LDDMM) \cite{durrleman2014morphometry} to establish point-to-point anatomical correspondence across anatomies through \textit{anatomical mapping}. 
This step ensures that vertices in corresponding regions of different meshes describe the same anatomical location, enabling subject-specific shapes to be compared and modeled in a common representation. 
In this representation, 
we represent each subject's anatomy by the initial momenta $\boldsymbol{\mu}_0 \in \mathbb{R}^{3 \times N_q}$, a vector field that defines how the template anatomy $\bar{\mathbf{\Gamma}} \in \mathbb{R}^{3 \times N}$ is deformed to match the subject-specific anatomy $\mathbf{\Gamma}_i$. 
Here, $N_q$ is a user-defined hyperparameter controlling the resolution of the momenta representation. Because $N_q$ is typically much smaller than the number of mesh vertices, the momenta provide a geometrically interpretable lower-dimensional representation of anatomical variability. 
We set $N_q = 720$, which empirically provided sufficient resolution for anatomical mapping, and used this representation as input to the autoencoder for further dimensionality reduction.

After generating synthetic momenta, we map the momenta-based shape representation back to 3D surface meshes through an inverse LDDMM process, also called \textit{geodesic shooting}.
This procedure deforms the template anatomy $\bar{\mathbf{\Gamma}}$ according to the generated momenta, resulting in a 3D biventricular surface mesh. 
We perform the LDDMM operations using the \textit{Deformetrica} software \cite{bone2018deformetrica}.
Details of the LDDMM formulation and the computational procedure used to convert images into shape representations are provided in~\ref{subsec:Appendix_LDDMM_details} and in prior works \cite{moscoloni2025unveiling, durrleman2014morphometry,bone2018deformetrica}.

\subsection{CAN-FLOW generative framework}
Figure~\ref{fig:fig_1} illustrates the proposed two-step generative strategy of CAN-FLOW, separating geometrical representation learning from conditional modeling of the anatomical distribution.

The first stage consisted of two encoding steps. 
We first represented each anatomy by its initial momenta, $\boldsymbol{\mu}_0 \in \mathbb{R}^{3 \times 720}$, using the lower-dimensional shape representation described in Section~\ref{sec:CMR-to-mesh}. 
We then trained an autoencoder to compress and reconstruct these momenta, yielding latent representations $\mathbf{z} \in \mathbb{R}^{44}$. 
Unlike cVAEs, \textit{the encoder and decoder were not explicitly conditioned on the metadata}, and the latent space was not constrained to follow a prescribed prior distribution. 
Therefore, the learned latent representations were derived only from geometry, while their metadata-dependent distribution was modeled separately by the conditional normalizing flow. 
We parametrized the encoder $\mathcal{E}_{\theta_{\mathcal{E}}}$ and decoder $\mathcal{D}_{\theta_{\mathcal{D}}}$, where $\theta_{\mathcal{E}}, \theta_{\mathcal{D}}$ were learnable parameters, with 3D convolutional layers, and used the Gaussian Error Linear Unit (GELU) activation function \cite{hendrycks2016gaussian}. 
We trained the autoencoder using mean squared error between the reference and reconstructed momenta and the Adam optimizer \cite{kingma2014adam}.

In the second stage of CAN-FLOW, we trained a conditional normalizing flow $f_{\phi}$ to model the metadata-dependent distribution of latent anatomical representations $\mathbf{z}$, where $\phi$ denoted learnable parameters.
For a given metadata vector $\mathbf{c} = [c_{sex}, c_{age}, c_{BMI}]$, the flow learned a bijective mapping from each latent representation $\mathbf{z}$ to a base variable $\mathbf{x} = f_{\phi}(\mathbf{z}, \mathbf{c})$.
The base variable was modeled with a metadata-dependent Gaussian distribution,
$\mathcal{N}(\mu_{\phi}(\mathbf{c}), \Sigma_{\phi}(\mathbf{c}))$, whose mean and covariance were predicted from $\mathbf{c}$ by small multi-layer perceptrons (MLPs).
After training, synthetic latent representations were generated by sampling a base variable
$\mathbf{x} \sim p(\mathbf{x}|\mathbf{c}) = \mathcal{N}(\mu_{\phi}(\mathbf{c}), \Sigma_{\phi}(\mathbf{c}))$
for the desired metadata $\mathbf{c}$ and applying the inverse flow,
$f^{-1}_{\phi}(\mathbf{x}, \mathbf{c}) = \tilde{\mathbf{z}}$.
A qualitative visualization of samples from the learned conditional base distribution was provided in Figure~\ref{fig:fig_appendix_5}.

We parameterized the bijective mapping as a composition of $N_b$ identical transformation blocks,
$f = f_\phi^{N_b} \circ f_\phi^{N_b-1} \circ \cdots \circ f_\phi^1.$
Our architecture was based on Glow \cite{kingma2018glow}, where each block consisted of an activation normalization layer, an invertible linear transformation, and an affine coupling layer. 
To condition the flow on metadata, we augmented each block with an affine injector layer and provided the metadata to the affine coupling layer.
The affine injector introduced a learned, metadata-dependent affine transformation that directly modulated the intermediate activations.
The metadata $\mathbf{c}$ were first embedded into a higher-dimensional representation $g(\mathbf{c}) \in \mathbb{R}^{12}$ using a small MLP, and this embedding was provided to both the affine injector and affine coupling layers.
We trained the conditional normalizing flow by minimizing the negative log-likelihood expressed as
\begin{equation}\label{eq:nflow_cond_loss_function}
    \mathcal{L}_{NLL}(\mathbf{z}, \mathbf{c}) = - \log \left(p_{X|C}(f_{\phi}(\mathbf{z, c})|\mathbf{c})\right) - \log \left(\left|\operatorname{det}\left(\frac{\partial f_{\phi}(\mathbf{z, c})}{\partial \mathbf{z}}\right)\right|\right),
\end{equation}

where $p_{X|C}(f_{\phi}(\mathbf{z, c})|\mathbf{c})$ denotes the conditional base density evaluated at the flow-transformed variable.
The second term of equation~\eqref{eq:nflow_cond_loss_function}  accounts for the change in volume induced by the transformation through 
the log-determinant of the Jacobian with respect to $\mathbf{z}$.
For the main experiments, we used $N_b=15$ identical normalizing-flow blocks and GELU activations throughout all neural components. Sensitivity to the number of blocks was reported in~\ref{sec:appendixAblationStudies}.
We trained the models with the Adam optimizer.
We developed all deep learning architectures with the PyTorch library \cite{paszke2019pytorch}, and trained them on an Intel Xeon Gold 6348 CPU.
More details on the architecture, training, and mathematical operations of normalizing the flow and the autoencoder were provided in~\ref{subsec:Appendix_autoencoder_architecture} and~\ref{subsec:Appendix_cnf_architecture}.

\subsection{cVAE baselines, training, and implementation}
We trained cVAE baselines using the same LDDMM-based momenta representation as CAN-FLOW. 
To ensure a fair comparison, the cVAE encoder and decoder were designed to match the architecture of the autoencoder used in CAN-FLOW, including the number of convolutional layers, hidden dimensions, activation functions, and latent dimensions. 
The key architectural difference was the latent shape representation conditioning mechanism: 
in the cVAE, metadata were provided to both the encoder and decoder, following the standard cVAE formulation \cite{sohn2015learning}.
During training, the encoder mapped each observed momenta representation $\boldsymbol{\mu}$ and its metadata $\mathbf{c}$ to an approximate posterior distribution in latent space. 
A latent sample $\mathbf{z}$ from this distribution was passed, together with $\mathbf{c}$, to the decoder to reconstruct $\boldsymbol{\mu}$. 
During generation, the encoder was not used: the model sampled a latent representation $\tilde{\mathbf{z}}$ directly from the shared, conditioning-agnostic Gaussian prior and combined it with the desired metadata $\mathbf{c}$ in the decoder to generate synthetic momenta $\tilde{\boldsymbol{\mu}}$ as
\begin{equation}
\label{eq:cvae_generation}
\tilde{\boldsymbol{\mu}} = \mathcal{D}_{\theta_{\mathcal{D}}}^{cVAE} (
\tilde{\mathbf{z}}, \mathbf{c}
),
\end{equation}
where $\mathcal{D}_{\theta_{\mathcal{D}}}^{cVAE}$ refers to the cVAE decoder. 
This differs from CAN-FLOW, where the conditional normalizing flow generates latent shape representations $\tilde{\mathbf{z}}$ conditioned on the metadata, which are then passed to a geometry-only decoder.

We trained the cVAE by maximizing the conditional evidence lower bound (ELBO), which approximated the data likelihood \cite{kingma2013auto}. 
Equivalently, we minimized the following negative ELBO loss:
\begin{equation}
\label{eq:cvae_loss}
\mathcal{L}_{\text{cVAE}}(\boldsymbol{\mu}, \mathbf{c};\beta)
= 
\underbrace{
\mathbb{E}_{q_{\theta_{\mathcal{E}}}(\mathbf{z} \mid \boldsymbol{\mu}, \mathbf{c})}
\left[
- \log p_{\theta_{\mathcal{D}}}(\boldsymbol{\mu} \mid \mathbf{z}, \mathbf{c})
\right]
}_{\text{reconstruction term}}
+
\beta \,
\underbrace{
\mathrm{KL}
\left(
q_{\theta_{\mathcal{E}}}(\mathbf{z} \mid \boldsymbol{\mu}, \mathbf{c})
\,\|\, 
p(\mathbf{z})
\right)
}_{\text{regularization term}}.
\end{equation}
Here, $q{_{\theta_{\mathcal{E}}}}(\mathbf{z} \mid \boldsymbol{\mu}, \mathbf{c})$ denotes the approximate posterior defined by the encoder, $p_{\theta_{\mathcal{D}}}(\boldsymbol{\mu} \mid \mathbf{z}, \mathbf{c})$ denotes the conditional likelihood defined by the decoder, and $p(\mathbf{z})$ is the conditioning-agnostic Gaussian prior. The encoder and decoder are parametrized by learnable weights $\theta_{\mathcal{E}}$ and $\theta_{\mathcal{D}}$, respectively.
This optimization differs from CAN-FLOW, which models the metadata-dependent density of the learned anatomical latent representations directly with a conditional normalizing flow rather than through a variational objective with a fixed Gaussian prior.

The KL regularization term encourages the latent samples supplied to the decoder during training to remain compatible with the Gaussian prior used for generation. 
Without this constraint, the decoder could reconstruct training anatomies accurately from encoder-produced latent codes, while samples drawn from the Gaussian prior might fall outside the regions of latent space encountered during training and decode into implausible anatomies. 
The parameter $\beta$ controls the trade-off between compatibility with prior-based generation and retaining anatomy-specific information in $\mathbf{z}$. 
Larger values strengthen the alignment with the prior but may reduce reconstruction accuracy by limiting the information encoded in $\mathbf{z}$. 
Smaller $\beta$ values place more emphasis on reconstruction fidelity but provide weaker alignment with the prior.
Since cVAE performance is sensitive to this trade-off \cite{rezende2018taming}, we train models with $\beta$ values spanning orders of magnitude from $10^{-6}$ to $10^{-1}$.
We train all cVAEs with the Adam optimizer.
Further details on the mathematical formulation of cVAEs and the derivation of the loss function are provided in \cite{kingma2013auto, sohn2015learning}.
Additional cVAE training details are provided in~\ref{subsec:Appendix_autoencoder_architecture}.

\clearpage
\section*{Data availability}
The UK Biobank data used in this study are available to approved researchers through standard UK Biobank access procedures at \hyperlink{https://www.ukbiobank.ac.uk/use-our-data/apply-for-access/}{https://www.ukbiobank.ac.uk/use-our-data/apply-for-access/}.
Derived data availability will be specified in accordance with UK Biobank requirements before publication.

\section*{Code availability}
The code used to train and evaluate CAN-FLOW and the cVAE baselines will be made available in a public repository at publication.

\section*{Acknowledgements}
This work was funded by the European Union’s Horizon
Europe research and innovation program (VITAL – Grant No.
101136728, to K.K., B.A., and M.P.), and Research Foundation – Flanders,
Fonds voor Wetenschappelijk Onderzoek – Vlaanderen (Grant
No. 11PS524N, to B.M.).
This work has been conducted using the UK Biobank Resource under Application Number 81032.

\section*{Author contributions}
Conceptualization: K.K., B.M., and M.P. 
Methodology: K.K., A.H., and M.P. 
Software, formal analysis and visualization: K.K. 
Data access, cohort definition and preprocessing: K.K., B.M., C.B., and J.A.C. 
Shape analysis and LDDMM processing: K.K. and B.M. 
Clinical interpretation: K.K., J.A.C., and M.P. 
Supervision: B.A., A.H., and M.P. 
Funding acquisition: M.P. 
Writing -- original draft: K.K. and M.P. 
Writing -- review and editing: K.K., B.M., B.A., J.A.C., A.H., and M.P.

\section*{Competing interests}
The authors declare no competing interests.

\clearpage
\appendix
\setcounter{figure}{0}
\section{Supplementary Methods}
\subsection{Diffeomorphic anatomical mapping and geodesic shooting}
\label{subsec:Appendix_LDDMM_details}
We employ LDDMM to achieve point-to-point correspondence and geometry-aware dimensionality-reduction, before the autoencoding step.
For each subject $i$, LDDMM represents the deformation by a set of $N_q$ control points $\{\mathbf{q}_j(t)\}_{j=1}^{N_q}$ and associated momenta $\{\boldsymbol{\mu}_{ij}(t)\}_{j=1}^{N_q}$, with $\mathbf{q}_j(t), \boldsymbol{\mu}_{ij}(t) \in \mathbb{R}^{3}$.
We set $N_q = 720$ because this control-point resolution provided sufficient anatomical mapping accuracy while keeping the momenta representation lower-dimensional than the original mesh \cite{moscoloni2025unveiling}.
To map the real surface meshes to the template anatomy $\mathbf{\bar{\Gamma}} \in \mathbb{R}^{3 \times N}$ and establish point-to-point correspondence, we optimize subject-specific momenta by minimizing the mean vertex-to-vertex error between the target anatomical meshes $\{\mathbf{\Gamma}_i\}_{i=1}^{N_p}$ and the template $\bar{\mathbf{\Gamma}}$ together with a deformation regularization term, where $N_p=2274$ is the number of anatomies.
The  loss functional reads \begin{equation}\label{eq:find_momenta_optim}
    \mathcal{L}(\mathbf{\bar{\Gamma}}, \boldsymbol{\mu}_i) = \sum_{i=1}^{N_p}\frac{1}{\sigma^2}d_W(\mathbf{\Phi}_{\mathbf{q}, \boldsymbol{\mu}_i}(\mathbf{\bar{\Gamma}}), \mathbf{\Gamma}_i)^2 + \sum_{i=1}^{N_p}\boldsymbol{\mu}_i^{\mathrm{T}}K_W(\mathbf{q}, \mathbf{q})\boldsymbol{\mu}_i.
\end{equation}
The first term measures the varifold distances $d_W$ between each target geometry $\mathbf{\Gamma}_i$ and the template $\bar{\mathbf{\Gamma}}$ \cite{charon2013varifold}.
The second term penalizes deformations with high kinetic energy, resulting in a unique set of momenta for each target geometry.
$\sigma$ acts as a weight, controlling the relative importance between the two summations.
$K_W$ is a Gaussian kernel $K_W(\mathbf{x}, \mathbf{y})=\exp(-\|\mathbf{x}-\mathbf{y}\|^2/\lambda_W)$, with kernel width $\lambda_W=10$.
The deformation of the template is defined by the diffeomorphism $\mathbf{\Phi}_{\mathbf{q}, \boldsymbol{\mu}_i}(\cdot): \mathbb{R}^3 \mapsto \mathbb{R}^{3}$.  
We compute the template geometry $\mathbf{\bar{\Gamma}}$ as the mean of the real patient-specific anatomies of our dataset.
The hyperparameter settings and initial template used for optimization follow the configuration used in \cite{moscoloni2025unveiling}.

After either CAN-FLOW or a cVAE baseline generates synthetic initial momenta $\tilde{\boldsymbol{\mu}}_0$, we use \textit{geodesic shooting} to map these momenta back to 3D surface meshes.
This inverse LDDMM step integrates the control-point and momenta dynamics over $t \in [0,1]$ 
and applies the resulting deformation to the template anatomy,
\begin{equation}
\label{eq:diffeq_momenta_cps}
    \left\{
    \begin{array}{l}
    \dot{\mathbf{q}}(t)=K_{V}(\mathbf{q}(t), \mathbf{q}(t)) \cdot \boldsymbol{\mu}(t), \\
    \dot{\boldsymbol{\mu}}(t)=-\frac{1}{2} \nabla_{\mathbf{q}}
    \left[
    K_{V}(\mathbf{q}(t), \mathbf{q}(t)) \cdot 
    \boldsymbol{\mu}(t)^{\mathrm{T}} \boldsymbol{\mu}(t)
    \right].
    \end{array}
    \right.
\end{equation}
$K_V$ is another Gaussian kernel of width $\lambda_V=10$. 
Given the synthetic initial momenta $\tilde{\boldsymbol{\mu}}_0$ and initial control points $\mathbf{q}(t_0=0)$, we integrate these equations to $t=1$. 
The resulting diffeomorphism deforms the template points $\mathbf{x} \in \mathbf{\bar{\Gamma}}$ into the synthetic surface mesh.

\subsection{Detection and removal of outlier anatomies}
\label{subsec:Appendix_OutlierRemoval}
After mesh extraction, visual inspection of the extracted biventricular shapes revealed a small number of unrealistically distorted anatomies. 
To remove such cases before training CAN-FLOW, we performed outlier detection in the space of momenta.
We first applied PCA to all momenta and retained the principal components explaining 99\% of the total variance. 
Each anatomy was then represented by its PCA score vector, and its deviation from the population was quantified using the Mahalanobis distance \cite{mahalanobis2018generalized},
\begin{equation}
D_M(\mathbf{s}) =
\sqrt{
(\mathbf{s}-\boldsymbol{\mu})^{T}
\mathbf{\Sigma}^{-1}
(\mathbf{s}-\boldsymbol{\mu})
},
\end{equation}
where $\mathbf{s}$ is the PCA score vector of an anatomy, $\boldsymbol{\mu}$ is the mean score vector, and $\mathbf{\Sigma}$ is the covariance matrix of the PCA scores.
Assuming that the PCA scores are approximately multivariate Gaussian, $D_M^2$ follows a chi-squared distribution with $k$ degrees of freedom, where $k$ is the number of retained principal components. 
We therefore classified an anatomy as an outlier if
\begin{equation}
D_M >
\sqrt{\chi^2_k(p)},
\end{equation}
with $p = 1-10^{-13}$.
This deliberately conservative threshold was chosen to remove only the most severe distortions, corresponding to samples located far from the main distribution in PCA space. 
This procedure identified 66 outliers, resulting in a final dataset of 2208 anatomies. 

\subsection{Autoencoder architecture and latent representation learning}
\label{subsec:Appendix_autoencoder_architecture}
We used a 44-dimensional latent space to encode the momenta representations, balancing reconstruction accuracy against latent representation size. 
Sensitivity to latent dimensionality is reported in~\ref{sec:appendixAblationStudiesLatentDim}.
Table~\ref{tab:tab1_appendix} lists the encoder layers, kernel sizes, strides, and output shapes.
To maintain architectural symmetry, the decoder closely follows the encoder, substituting convolutional layers with transposed 3D convolutions to invert the encoding process.
We do not apply any activation function to the decoder output.  
We trained the autoencoder and cVAEs for 2000 epochs using Adam, a learning rate of $2 \cdot 10^{-4}$, and a batch size equal to 15\% of the training dataset. 
We applied an $\ell_2$ penalty with regularization strength $\lambda=10^{-4}$ during autoencoder and cVAE training.
The developed cVAEs used for the main experiments comprise 6,392,227 learnable parameters.

\begin{table}[h!]
\caption{\textbf{The technical details of the layers that comprise the encoder part of the autoencoder.} Layer types, kernel sizes, strides, and activation functions are listed. The decoder follows a symmetric architecture with transpose convolutions, without the last GELU activation.}
\begin{tabular}{c|ccccc}
\hline
\toprule
\textbf{Layer} & \textbf{Input channels} & \textbf{Output channels} & \textbf{Kernel size} & \textbf{Stride} & \textbf{Output shape}  \\ \midrule
3D conv         & 3                       & 32                       & (1, 1, 1)            & 1               & (32, 8, 9, 10)      \\ 
GELU activation&                         &                      &                      &                 &                       \\ 
\hline
3D conv         & 32                      & 32                       & (2, 2, 2)            & 1               & (32, 7, 8, 9)         \\ 
GELU activation &                         &                        &                      &                 &                       \\ \hline
3D conv         & 32                      & 64                       & (1, 1, 1)            & 1               & (64, 7, 8, 9)         \\
GELU activation &                         &                        &                      &                 &                       \\ \hline
3D conv         & 64                      & 128                       & (2, 2, 2)            & 1               & (128, 6, 7, 8)         \\  GELU activation &                         &                        &                      &                 &                       \\ \hline
Linear layer & $128\cdot6\cdot7\cdot8$ & 44 & $-$ & $-$ & $(44)$ \\
\bottomrule
\end{tabular}
\label{tab:tab1_appendix}
\end{table}

\subsection{Conditional normalizing-flow architecture}
\label{subsec:Appendix_cnf_architecture}
We implemented the conditional normalizing flow using Glow-style blocks composed of activation normalization, an invertible linear transformation, and an affine coupling layer \cite{dinh2014nice,dinh2016density, kingma2018glow, ho2019flowplusplus}. 
Metadata conditioning was introduced through the affine coupling layers, an affine injector layer, and a conditional base distribution.

\textbf{Conditional affine coupling.}
Following the conditioning strategy of \cite{winkler2019learning}, we modify the Glow affine coupling layer to incorporate the metadata vector $\mathbf{c}$. Given an input split into two parts, $\mathbf{b}_0$ and $\mathbf{b}_1$, the conditional affine coupling transformation is defined as
\begin{equation}\label{eq:affinecoupling_cond_fwd}
\mathbf{h}_0^{mlp} = s(\mathbf{b}_1, g(\mathbf{c}))\cdot \mathbf{b}_0 + t(\mathbf{b}_1, g(\mathbf{c})),
\quad
\mathbf{h}_1^{mlp} = \mathbf{b}_1,
\end{equation}
where $s(\cdot)$ and $t(\cdot)$ denote the scale and translation functions, respectively. Both functions are parametrized as small MLPs. The function $g(\cdot)$ is another MLP that maps the metadata vector $\mathbf{c}$ to a higher-dimensional embedding, allowing the coupling layer to use a richer representation of the conditioning information.

\textbf{Affine injector.}
We added an affine injector layer inspired by \cite{lugmayr2020srflow} to apply a metadata-dependent affine transformation to the full input $\mathbf{b}$:
\begin{equation}\label{eq:affine_injector}
\mathbf{h}^{a.i.} = s(\mathbf{c}) \cdot \mathbf{b} + t(\mathbf{c}),
\end{equation}
where $\mathbf{h}^{a.i.}$ is the output of the affine injector layer. Here, $s(\mathbf{c})$ and $t(\mathbf{c})$ are the scale and translation functions of the affine injector. 
These are again small MLPs that depend only on the metadata. 
In contrast to affine coupling layers, the affine injector does not split the input into two branches. As a result, all elements of $\mathbf{b}$ are transformed directly as a function of the metadata.

\textbf{Conditional prior.}
In addition to conditioning the flow transformations, we also utilize a metadata-dependent base distribution. Instead of adopting a fixed standard normal prior, as in non-conditional normalizing flows \cite{dinh2014nice, dinh2016density, kingma2018glow}, we define the base density as
\begin{equation}\label{eq:conditional_prior}
p(\mathbf{x}\mid\mathbf{c}) = \mathcal{N}(\mathbf{x};\mu(\mathbf{c}), \boldsymbol{\Sigma}(\mathbf{c})),
\end{equation}
where the mean $\mu(\mathbf{c})$ and covariance $\boldsymbol{\Sigma}(\mathbf{c})$ are parametrized by small MLPs. 
This allows sampling to depend on metadata at the base-distribution level before applying the inverse flow.
Consequently, different regions of the base distribution can be associated with different metadata profiles, enabling controllable generation of synthetic anatomies conditioned on $\mathbf{c}$.

\begin{figure}[h!]
\centering
  \includegraphics[width=\linewidth]{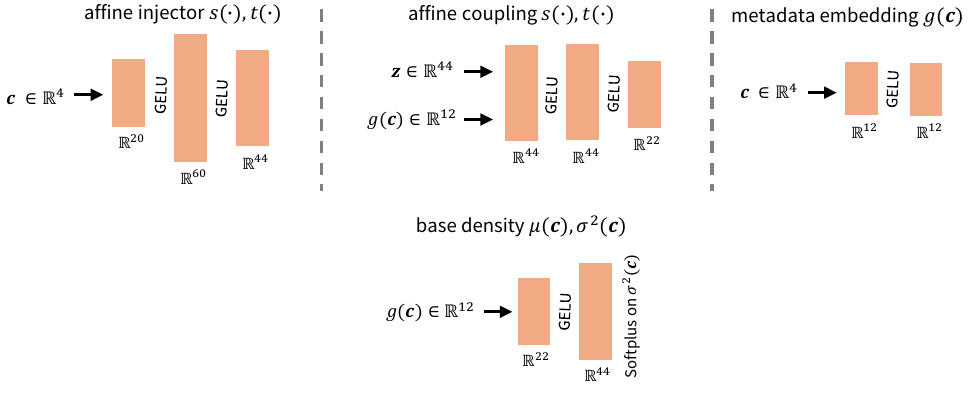}
  \caption{\textbf{Architectures of the shallow MLPs used in CAN-FLOW.}
  All models have at most a single hidden layer.
  An orange block denotes a fully connected neural network layer, whose width is reported below the block. 
  Activation functions are reported between the blocks.
  Consistent with the paper notation, latent representations are denoted with $\mathbf{z}$, metadata are denoted with $\mathbf{c}$, and $g(\mathbf{c})$ refers to the metadata embedding.}
  \label{fig:fig_appendix_1}
\end{figure}

The architectures of the small MLPs used for the scale and translation functions, the metadata embedding, and the conditional base distribution are reported in Figure~\ref{fig:fig_appendix_1}.

Our main CAN-FLOW model used $N_b = 15$ identical flow blocks.
Before training, we initialize the parameters of the affine injector and affine coupling transformation to zero, as this helps accelerate convergence of the loss function.
Similarly to the autoencoder training, we impose a $\ell_2$ regularization penalty on the weights of the normalizing flow, with regularization weight $\lambda=10^{-5}$. We train the framework for 1200 epochs with a learning rate of $2.5 \cdot 10^{-4}$.
Sex was one-hot-encoded as $[1,0]$ for female subjects and $[0,1]$ for male subjects.
We then scale all features, including age, BMI, and the encoded sex components, to the range $[0, 1]$ to facilitate learning, with the MinMaxScaler routine of sklearn \cite{pedregosa2011scikit}. 
The developed normalizing flow used for the main experiments comprises 4,299,379 learnable parameters.

\subsection{Phenotype computation and model evaluation}\label{subsec:Appendix_performance_metrics}
For all evaluations, real and synthetic anatomies were compared using the same metadata sampling strategy. 
Age and BMI were sampled uniformly from the ranges observed in the real cohort, separately for female and male subjects. 
The overall-population synthetic cohorts contained 600 anatomies, balanced by sex with 300 female and 300 male samples.

\subsubsection{Phenotype computation from biventricular meshes}
For each real and synthetic anatomy, we computed LVEDV, RVEDV, myocardial mass, RVEDV-to-LVEDV ratio, LV long-axis length, and LV sphericity.
All computations were performed on subject-specific surface meshes exported in VTK format using PyVista \cite{sullivan2019pyvista}.

Ventricular cavity volumes were computed from closed LV and RV endocardial surfaces, while
myocardial volume was computed as the total epicardial volume minus the combined LV and RV cavity volumes.
Myocardial mass was obtained by multiplying the myocardial volume by a tissue density of 1.05~g/mL.
The LV long-axis length was computed as the distance between the apex and the center of the mitral valve opening. 
The mitral valve opening was identified as the largest open boundary of the isolated LV mesh.  
The apex was then defined as the LV surface point farthest from the plane fitted to this opening.
LV sphericity was defined as the LV endocardial volume normalized by the volume of a sphere with the same diameter as the LV long-axis length. This provides a dimensionless measure of how spherical the LV cavity is, with higher values corresponding to a more spherical shape and lower values corresponding to a more elongated shape.
LV sphericity and long-axis length provide clinically meaningful descriptors of ventricular remodeling, as they quantify the transition from the normal elongated LV geometry toward a more spherical shape. This geometric remodeling has been shown to provide prognostic information for ventricular tachyarrhythmia risk in patients with LV systolic dysfunction, beyond conventional measures such as ejection fraction alone \cite{nakamori2017left, medrano2014left}.

\subsubsection{Point-cloud coverage and MMD computation}
To compute MMD and point-cloud coverage, we transform real and synthetic surface meshes into point clouds, and quantify point cloud similarity in terms of Chamfer distance (CD) as follows: 
\begin{equation}
\mathrm{CD}(X, Y) 
= \sum_{x \in X} \min_{y \in Y} \| x - y \|_2^2 
+ \sum_{y \in Y} \min_{x \in X} \| x - y \|_2^2,
\end{equation}
where X and Y denote two point clouds. 
Given a generated ($S_g$) and a real ($S_r$) cohort of point clouds, we define MMD as: 
\begin{equation}
    \mathrm{MMD}(S_g, S_r) = \frac{1}{|S_r|}\sum_{Y \in S_r} \min_{X \in S_g} CD(X, Y),
\end{equation}
where $|S_r|$ denotes the number of real point clouds. 
In this work, MMD quantifies geometric fidelity by measuring, for each real anatomy, the Chamfer distance to its nearest synthetic anatomy.
Lower values indicate that the generated shapes exhibit smaller structural discrepancies relative to the real cohort.
We compute the coverage score as the proportion of real anatomies that are matched to at least one synthetic anatomy based on CD similarity. 
This metric evaluates how well the synthetic cohort captures the anatomical variability of the real cohort, and is defined as:
\begin{equation}
    \mathrm{COV}(S_g, S_r) = \frac{\left|\{\arg \min_{Y \in S_r} CD(X, Y)|X \in S_g\}\right|}{|S_r|}.
\end{equation}
Higher values indicate a synthetic cohort that captures a broader range of the real anatomical variability.

\subsubsection{Wasserstein distance and KL divergence computation}
We compute the distributional similarity between real and synthetic distributions using two complementary metrics, Wasserstein distance and KL divergence. 
The Wasserstein distance was computed directly from the samples using the corresponding scipy routine \cite{2020SciPy-NMeth}.
Regarding KL divergence, the reference and generated distributions were estimated using normalized histograms with 25 equally spaced bins. 
We discretize each distribution using 25 shared histogram bins, providing a balance between resolution and ensuring that each bin contains a sufficient number of samples.
We define common bin edges over the combined range of both datasets to ensure direct comparability. 
We then compute the KL divergence as
\begin{equation}
   D_{\mathrm{KL}}(P \| Q) = \sum_{b=1}^{25} p_b
\log\left(\frac{p_b}{q_b}\right), 
\end{equation}
where $P$ denotes the real distribution and $Q$ the synthetic distribution.

\subsubsection{PCA-based variability analysis in momenta space}
\label{subsubsec:AppendixPCAEXPMETHODS}
We used PCA to quantify anatomical variability directly in momenta space, capturing shape differences that may not be reflected in the mesh-derived clinical phenotypes alone.
We randomly split the cohort of real momenta into two subsets of equal size $\mathbf{Y}_1, \mathbf{Y}_2 \in \mathbb{R}^{(3N_q) \times N_Y}$. 
We perform PCA on $\mathbf{Y}_1$, and project both $\mathbf{Y}_1$ and $\mathbf{Y}_2$ onto the computed PCA basis modes $\mathbf{U}$, to obtain the PCA coefficients:
\begin{equation}
    \mathbf{X}_1 = \mathbf{U}^{\mathrm{T}}\mathbf{Y}_1, \space \mathbf{X}_2 = \mathbf{U}^{\mathrm{T}}\mathbf{Y}_2.
\end{equation}
For each retained PCA coefficient, we measured the Wasserstein distance between the coefficient distributions of the two subsets:
\begin{equation}\label{eq:pca_experiment_nominator}
    \mathrm{WD}(\mathbf{X}_1, \mathbf{X}_2).
\end{equation}
By measuring the distributional difference between the principal components $\mathbf{X}_1$ and $\mathbf{X}_2$, we quantify how different these two real cohort subsets are.
As a result, we use Equation~\eqref{eq:pca_experiment_nominator} as a representation of the anatomical variability of the real cohort.
Then, we substitute one of the real subsets with sets of synthetic anatomies $\mathbf{\tilde{Y}}_{\text{gen}}$ from different generative models, and compute the Wasserstein distance between the distributions of coefficients, defined as 
\begin{equation}\label{eq:pca_experiment_denominator}
    \mathrm{WD}(\mathbf{X}_1, \mathbf{\tilde{X}}_{\text{gen}}),
\end{equation}
where $\mathbf{\tilde{X}}_{\text{gen}} = \mathbf{U}^{\mathrm{T}}\mathbf{\tilde{Y}}_{\text{gen}}$.
Similarity between synthetic and real cohorts was quantified as:
\begin{equation}
    r = \frac{\mathrm{WD}(\mathbf{X}_1, \mathbf{X}_2)}{\mathrm{WD}(\mathbf{X}_1, \tilde{\mathbf{X}}_{\text{gen}})}.
\end{equation}
A generated cohort is considered similar to the real cohort when this ratio for most PCA coefficients approaches $r \approx 1$, indicating faithful reproduction of anatomical variability. 
To account for the stochasticity in the split of [$\mathbf{Y}_1$, $\mathbf{Y}_2$], we perform the experiment with five random splits of the real subset, and visualize the confidence interval of one standard deviation.

To investigate whether the generative models can reproduce the sex-specific variability of the real cohort, we split the real cohort in female and male subsets $\mathbf{Y}^f$ and $\mathbf{Y}^m$. 
We perform PCA on the male subset $\mathbf{Y}^m$, and then substitute the female subset with synthetic female cohorts from different generative models. 
Analogously, we define the sex-specific ratio as:
\begin{equation}
    r = \frac{\mathrm{WD}(\mathbf{X}^m, \mathbf{X}^f)}{\mathrm{WD}(\mathbf{X}^m, \tilde{\mathbf{X}}^f_{\text{gen}})},
\end{equation}
where $\mathbf{X}^m = \mathbf{U}^\mathrm{T}\mathbf{Y}^m$, $\mathbf{X}^f = \mathbf{U}^\mathrm{T}\mathbf{Y}^f$.
The same interpretation of $r$ applies to this sex-specific comparison.

\clearpage
\section{Supplementary Results}
\label{sec:AppendixSupplementaryResults}
\subsection{Additional cVAE regularization results}
\label{subsec:Appendix_additional_cvae_results}
In the main manuscript, we report the two strongest cVAE configurations, $\beta=10^{-2}$ and $\beta=10^{-3}$. 
Here, we show the remaining cVAE baselines trained over the predefined range $\beta \in [10^{-6}, 10^{-1}]$, which is consistent with prior conditional cardiac generative modeling studies \cite{qiao2023cheart, Qiao2025, beetz2022interpretable, beetz2022multi}.

\begin{figure}[h!]
\centering
  \includegraphics[width=\linewidth]{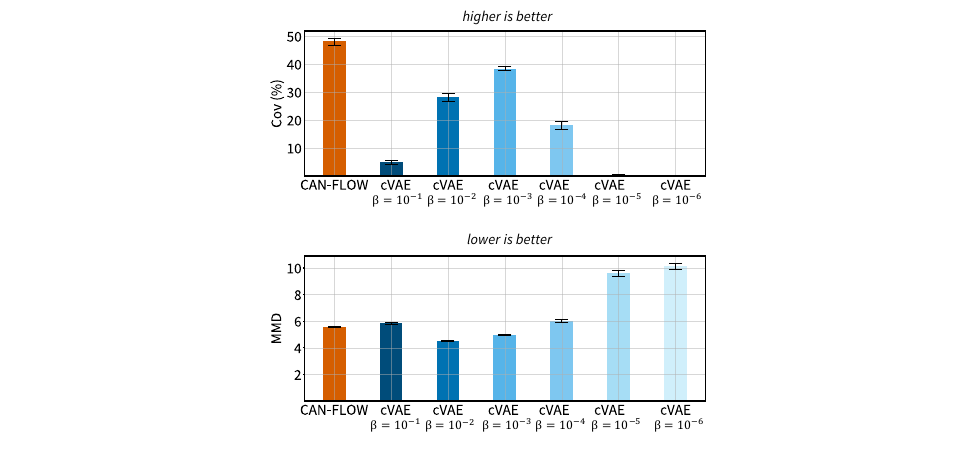}
    \caption{\textbf{Comparison of the generative model performance in terms of MMD and coverage metrics.} 
    These shape-based metrics are computed in terms of Chamfer distance. 
    To keep $|S_r| = |S_g|$, 600 anatomies were subsampled from the real cohort, and the mean and standard deviation are reported over four subsampling repetitions. 
    CAN-FLOW is compared with cVAEs trained with varying levels of regularization.}
  \label{fig:fig_appendix_2}
\end{figure}

Figure~\ref{fig:fig_appendix_2} illustrates the coverage and MMD metrics for CAN-FLOW and all evaluated cVAE configurations. 
Within this range, both the strongest regularization setting ($\beta=10^{-1}$) and the weakest regularization settings ($\beta<10^{-3}$) showed worse coverage and MMD values than the cVAE configurations reported in the main text.

\begin{figure}[h!]
\centering
  \includegraphics[width=\linewidth]{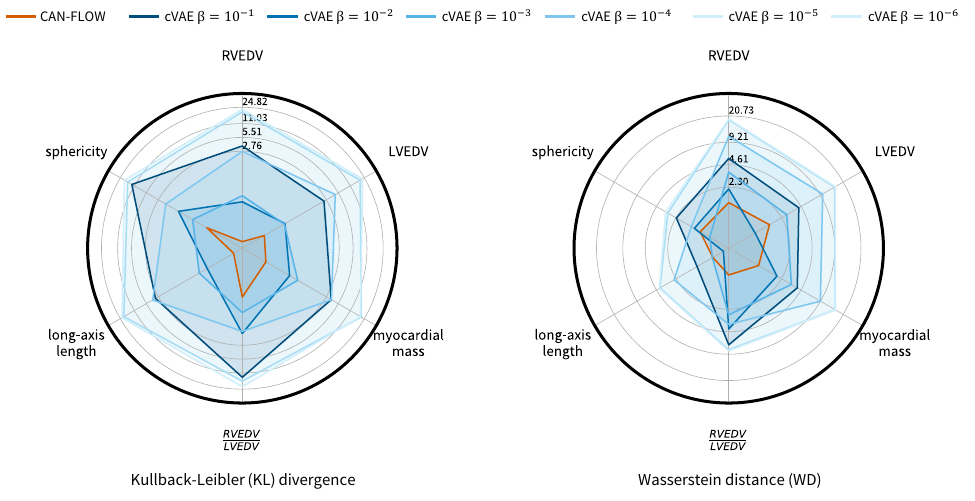}
    \caption{\textbf{Comparison of the generative performance of CAN-FLOW and cVAEs, in terms of KL divergence and Wasserstein distance metrics.} 
    The radar plots compare the real and synthetic distributions of clinical phenotypes. 
    A smaller metric value corresponds to better similarity between the generated and real phenotype distributions.
    CAN-FLOW is compared with all cVAEs trained in this study. 
    The radial axis is displayed on a logarithmic scale.}
  \label{fig:fig_appendix_3}
\end{figure}

In Figure~\ref{fig:fig_appendix_3}, the clinical phenotype metrics showed the same pattern: cVAE configurations outside the two main-text settings produced larger Wasserstein distance and KL divergence values across most phenotypes.
This further illustrates the sensitivity of cVAE performance to the choice of regularization strength.

\subsection{Additional PCA experiment results}
\label{subsec:appendixPCAexperiment}
\begin{figure}[h!]
\centering
  \includegraphics[width=\linewidth]{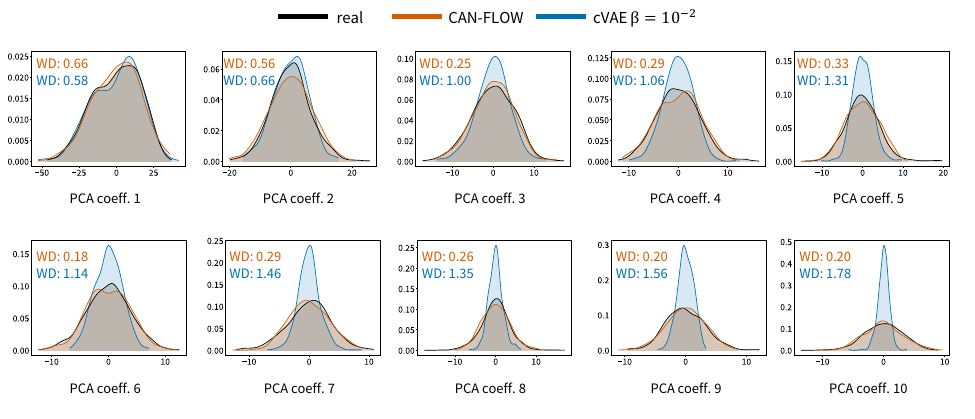}
  \caption{\textbf{Marginal distributions of the leading PCA coefficients in shape momenta space.} 
  The plots compare the real PCA shape coefficient distributions with those obtained from CAN-FLOW-generated cohorts and the most competitive cVAE baseline in terms of cohort diversity. 
  Densities are estimated using kernel density estimation. 
  Wasserstein distance (WD) values quantify the discrepancy between each synthetic distribution and the corresponding real distribution for CAN-FLOW (orange) and cVAE with $\beta=10^{-2}$ (blue).
  Lower values indicate closer agreement. 
  Better agreement across these coefficients indicates better preservation of the main modes of anatomical variability.}
  \label{fig:fig_appendix_4}
\end{figure}

Figure~\ref{fig:fig_appendix_4} shows the marginal distributions of the ten leading PCA coefficients from the momenta-space PCA analysis.
These marginal distributions provide a coefficient-wise view of the PCA-based variability comparison reported in the main text.
For most PCA coefficients, CAN-FLOW more closely follows the empirical real-cohort density than the most competitive cVAE, particularly in the distribution tails.
The Wasserstein distances support this visual pattern, with CAN-FLOW achieving the smaller distance for nine of the ten displayed coefficients. 
The cVAE distributions are more concentrated around the mean for several coefficients, consistent with the reduced variability observed in the main PCA analysis.

\subsection{Learned conditional base density}
\label{sec:appendixNFBaseDensity}
\begin{figure}[h!]
\centering
  \includegraphics[width=0.55\linewidth]{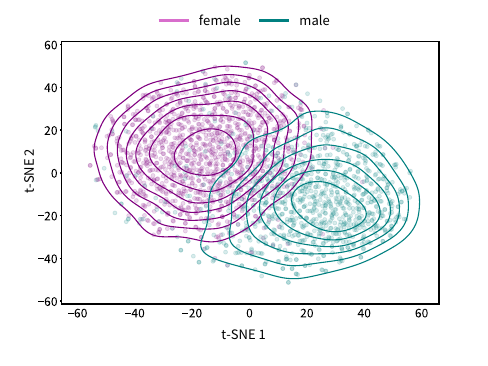}
  \caption{\textbf{The learned conditional base density of CAN-FLOW.}
  Female and male anatomy representations are shown in different colors. 
  The two-dimensional embedding was computed using t-SNE, and densities were estimated using kernel density estimation.}
  \label{fig:fig_appendix_5}
\end{figure}

Figure~\ref{fig:fig_appendix_5} visualizes samples from the learned conditional base distribution of CAN-FLOW in a reduced-dimensional t-SNE space \cite{van2008visualizing}. 
The sex-specific separation in this visualization is consistent with metadata-dependent structure in the learned base distribution, in contrast to the shared conditioning-agnostic Gaussian prior used in the cVAE baselines.

\subsection{CAN-FLOW architectural ablation study}
\label{sec:appendixAblationStudies}
We investigated the sensitivity of CAN-FLOW performance to selected hyperparameters of the normalizing-flow architecture.
In particular, we focused on three architectural factors with direct influence on flow capacity and conditioning: activation function, number of transformation blocks, and dimensionality of the metadata embedding $g(\mathbf{c})$.
These parameters were varied because they appeared to have the greatest effect on model performance during preliminary experimentation.
For every configuration, we train the conditional normalizing flow three times to account for the random weight initialization.
For each model we report the Wasserstein distance and KL divergence between the real and synthetic distributions of clinical phenotypes. 
We compare the different CAN-FLOW architectures with the best cVAE performance achieved for each evaluation metric in the Results section.
The results of the ablation study are reported in Tables~\ref{tab:tab2_appendix},~\ref{tab:tab3_appendix}, and~\ref{tab:tab4_appendix}.
Across the three ablation settings, the worst CAN-FLOW value in each KL-divergence column was lower than the corresponding best cVAE value in 16 of 18 comparisons. 
For Wasserstein distance, the worst CAN-FLOW value remained lower than the corresponding cVAE value in 8 of 18 comparisons.
These results support the interpretation that CAN-FLOW's performance does not depend on a highly optimized normalizing-flow architecture, since competitive results were obtained across several architectural choices, including models with fewer flow blocks.
\begin{table*}[h!]
    \centering
    \caption{\textbf{Comparison of the CAN-FLOW generative performance under different normalizing flow architectures in terms of the activation function used.} 
    KL divergence and Wasserstein distance are reported between real and synthetic clinical phenotype distributions. For each configuration, values indicate the mean and standard deviation over three runs with different random initializations. The largest value in each column is marked in \textcolor{blue}{blue}.}
    \label{tab:tab2_appendix}
    
    \renewcommand{\arraystretch}{1.15}
    \setlength{\tabcolsep}{5pt}
    
    \resizebox{\textwidth}{!}{
    \begin{tabular}{l|cccccc}
        \toprule
        \textbf{Activation function}
        & \multicolumn{6}{c}{
            \textbf{Kullback--Leibler (KL) divergence ($\downarrow$)}
        } \\
        \cmidrule(lr){2-7}
        & \textbf{LVEDV}
        & \textbf{RVEDV}
        & \makecell{\textbf{Myocardial}\\\textbf{mass}}
        & \makecell{\textbf{RVEDV/LVEDV}\\\textbf{ratio}}
        & \makecell{\textbf{Long-axis}\\\textbf{length}}
        & \makecell{\textbf{LV}\\\textbf{sphericity}} \\
        \midrule
        
        ELU~\cite{clevert2020fast}
        & $0.098 \pm 0.011$
        & $0.070 \pm 0.010$
        & $0.058 \pm 0.013$
        & $0.196 \pm 0.064$
        & \textcolor{blue}{$0.167 \pm 0.051$}
        & $0.144 \pm 0.061$ \\
        
        Leaky ReLU~\cite{xu2015empirical}
        & $0.085 \pm 0.011$
        & $0.076 \pm 0.040$
        & $0.054 \pm 0.029$
        & $0.221 \pm 0.082$
        & $0.047 \pm 0.007$
        & $0.281 \pm 0.018$ \\
        
        ReLU~\cite{agarap2018deep}
        & $0.100 \pm 0.036$
        & $0.093 \pm 0.042$
        & $0.071 \pm 0.013$
        & $0.178 \pm 0.061$
        & $0.098\pm 0.038$
        & $0.152 \pm 0.091$ \\
        
        SiLU~\cite{elfwing2018sigmoid}
        & $0.129 \pm 0.051$
        & $0.066 \pm 0.021$
        & $0.067 \pm 0.005$
        & $0.172 \pm 0.101$
        & $0.149 \pm 0.106$
        & $0.068 \pm 0.037$ \\
        
        cVAE
        & \textcolor{blue}{$0.244$}
        & \textcolor{blue}{$0.219$}
        & \textcolor{blue}{$0.331$}
        & \textcolor{blue}{$0.537$}
        & $0.164$
        & \textcolor{blue}{$0.379$} \\

        \midrule
        \addlinespace[0.25em]
        
        \textbf{Activation function}
        & \multicolumn{6}{c}{
            \textbf{Wasserstein distance (WD) ($\downarrow$)}
        } \\
        \cmidrule(lr){2-7}
        & \textbf{LVEDV}
        & \textbf{RVEDV}
        & \makecell{\textbf{Myocardial}\\\textbf{mass}}
        & \makecell{\textbf{RVEDV/LVEDV}\\\textbf{ratio}}
        & \makecell{\textbf{Long-axis}\\\textbf{length}}
        & \makecell{\textbf{LV}\\\textbf{sphericity}} \\
        \midrule
        
        ELU~\cite{clevert2020fast}
        & $1.277 \pm 0.071$
        & $1.340 \pm 0.262$
        & $1.267 \pm 0.118$
        & $1.303 \pm 0.269$
        & $0.303 \pm 0.011$
        & $0.490 \pm 0.055$ \\
        
        Leaky ReLU~\cite{xu2015empirical}
        & \textcolor{blue}{$1.481 \pm 0.683$}
        & $2.137 \pm 1.023$
        & $0.847 \pm 0.061$
        & $1.697 \pm 0.279$
        & \textcolor{blue}{$0.520 \pm 0.182$}
        & \textcolor{blue}{$0.840 \pm 0.139$} \\
        
        ReLU~\cite{agarap2018deep}
        & $1.274 \pm 0.503$
        & $1.110\pm 0.290$
        & $1.176 \pm 0.267$
        & $1.074 \pm 0.751$
        & $0.504 \pm 0.222$
        & $0.737 \pm 0.130$ \\
        
        SiLU~\cite{elfwing2018sigmoid}
        & $0.923 \pm 0.217$
        & $1.720 \pm 0.334$
        & $1.304 \pm 0.370$
        & $1.193 \pm 0.507$
        & $0.525 \pm 0.153$
        & $0.684 \pm 0.172$ \\
        
        cVAE
        & $0.899$
        & \textcolor{blue}{$2.184$}
        & \textcolor{blue}{$1.958$}
        & \textcolor{blue}{$2.734$}
        & $0.425$
        & $0.690$ \\
        
        \bottomrule
    \end{tabular}
    }
\end{table*}
\begin{table*}[h!]
    \centering
    \caption{\textbf{Comparison of the CAN-FLOW generative performance under different normalizing flow architectures in terms of the number of blocks.} 
    KL divergence and Wasserstein distance are reported between real and synthetic clinical phenotype distributions. For each configuration, values indicate the mean and standard deviation over three runs with different random initializations. The largest value in each column is marked in \textcolor{blue}{blue}.}
    \label{tab:tab3_appendix}
    
    \renewcommand{\arraystretch}{1.15}
    \setlength{\tabcolsep}{5pt}
    
    \resizebox{\textwidth}{!}{
    \begin{tabular}{l|cccccc}
        \toprule
        \textbf{\shortstack{Number\\of blocks} }
        & \multicolumn{6}{c}{
            \textbf{Kullback--Leibler (KL) divergence ($\downarrow$)}
        } \\
        \cmidrule(lr){2-7}
        & \textbf{LVEDV}
        & \textbf{RVEDV}
        & \makecell{\textbf{Myocardial}\\\textbf{mass}}
        & \makecell{\textbf{RVEDV/LVEDV}\\\textbf{ratio}}
        & \makecell{\textbf{Long-axis}\\\textbf{length}}
        & \makecell{\textbf{LV}\\\textbf{sphericity}} \\
        \midrule
        
        5
        & $0.072 \pm 0.007$
        & $0.110 \pm 0.031$
        & $0.070 \pm 0.031$
        & $0.213 \pm 0.078$
        & $0.060 \pm 0.039$
        & $0.122 \pm 0.017$ \\
        
        10
        & $0.105 \pm 0.018$
        & $0.049 \pm 0.017$
        & $0.055 \pm 0.021$
        & $0.133 \pm 0.008$
        & $0.125 \pm 0.071$
        & $0.327 \pm 0.149$ \\
        
        20
        & $0.098 \pm 0.018$
        & $0.073 \pm 0.016$
        & $0.092 \pm 0.028$
        & $0.181 \pm 0.057$
        & $0.126 \pm 0.083$
        & $0.147 \pm 0.061$ \\
        
        25
        & $0.067 \pm 0.011$
        & $0.089 \pm 0.068$
        & $0.082 \pm 0.029$
        & $0.224 \pm 0.056$
        & $0.059 \pm 0.036$
        & $0.136 \pm 0.071$ \\

        30
        & $0.091 \pm 0.012$
        & $0.095 \pm 0.026$
        & $0.146 \pm 0.044$
        & $0.135 \pm 0.056$
        & $0.061 \pm 0.044$
        & $0.146 \pm 0.008$ \\
        
        cVAE
        & \textcolor{blue}{$0.244$}
        & \textcolor{blue}{$0.219$}
        & \textcolor{blue}{$0.331$}
        & \textcolor{blue}{$0.537$}
        & \textcolor{blue}{$0.164$}
        & \textcolor{blue}{$0.379$} \\

        \midrule
        \addlinespace[0.25em]
        
        \textbf{\shortstack{Number\\of blocks} }
        & \multicolumn{6}{c}{
            \textbf{Wasserstein distance (WD) ($\downarrow$)}
        } \\
        \cmidrule(lr){2-7}
        & \textbf{LVEDV}
        & \textbf{RVEDV}
        & \makecell{\textbf{Myocardial}\\\textbf{mass}}
        & \makecell{\textbf{RVEDV/LVEDV}\\\textbf{ratio}}
        & \makecell{\textbf{Long-axis}\\\textbf{length}}
        & \makecell{\textbf{LV}\\\textbf{sphericity}} \\
        \midrule
        
        5
        & $1.119 \pm 0.247$
        & $1.170 \pm 0.385$
        & $0.928 \pm 0.293$
        & $1.344 \pm 0.119$
        & $0.367 \pm 0.114$
        & $0.652 \pm 0.130$ \\
        
        10
        & \textcolor{blue}{$1.261 \pm 0.373$}
        & $1.546 \pm 0.258$
        & $1.332 \pm 0.201$
        & $1.035 \pm 0.175$
        & $0.555 \pm 0.059$
        & \textcolor{blue}{$0.813 \pm 0.148$} \\
        
        20
        & $1.229 \pm 0.402$
        & $1.558\pm 0.205$
        & $0.736 \pm 0.108$
        & $1.023 \pm 0.126$
        & $0.455 \pm 0.140$
        & $0.609 \pm 0.026$ \\
        
        25
        & $1.110 \pm 0.215$
        & $0.954 \pm 0.091$
        & $0.980 \pm 0.119$
        & $1.066 \pm 0.299$
        & $0.453 \pm 0.088$
        & $0.639 \pm 0.182$ \\

        30
        & $1.234 \pm 0.218$
        & $1.627 \pm 0.220$
        & $1.497 \pm 0.326$
        & $0.723 \pm 0.099$
        & \textcolor{blue}{$0.571 \pm 0.102$}
        & $0.684 \pm 0.054$ \\
        
        cVAE
        & $0.899$
        & \textcolor{blue}{$2.184$}
        & \textcolor{blue}{$1.958$}
        & \textcolor{blue}{$2.734$}
        & $0.425$
        & $0.690$ \\
        
        \bottomrule
    \end{tabular}
    }
\end{table*}
\begin{table*}[h!]
    \centering
    \caption{\textbf{Comparison of the CAN-FLOW generative performance under different normalizing flow architectures in terms of the dimensionality of the metadata embedding $g(\mathbf{c})$.} 
    KL divergence and Wasserstein distance are reported between real and synthetic clinical phenotype distributions. For each configuration, values indicate the mean and standard deviation over three runs with different random initializations. The largest value in each column is marked in \textcolor{blue}{blue}.}
    \label{tab:tab4_appendix}
    
    \renewcommand{\arraystretch}{1.15}
    \setlength{\tabcolsep}{5pt}
    
    \resizebox{\textwidth}{!}{
    \begin{tabular}{l|cccccc}
        \toprule
        \textbf{\shortstack{$g(\mathbf{c})$: Embedding\\ dimension} }
        & \multicolumn{6}{c}{
            \textbf{Kullback--Leibler (KL) divergence ($\downarrow$)}
        } \\
        \cmidrule(lr){2-7}
        & \textbf{LVEDV}
        & \textbf{RVEDV}
        & \makecell{\textbf{Myocardial}\\\textbf{mass}}
        & \makecell{\textbf{RVEDV/LVEDV}\\\textbf{ratio}}
        & \makecell{\textbf{Long-axis}\\\textbf{length}}
        & \makecell{\textbf{LV}\\\textbf{sphericity}} \\
        \midrule
        
        6
        & $0.080 \pm 0.002$
        & $0.073 \pm 0.016$
        & $0.058 \pm 0.035$
        & $0.161 \pm 0.056$
        & $0.108 \pm 0.098$
        & $0.121 \pm 0.062$ \\
        
        9
        & $0.114 \pm 0.030$
        & $0.068 \pm 0.020$
        & $0.068 \pm 0.024$
        & $0.197 \pm 0.092$
        & \textcolor{blue}{$0.212 \pm 0.130$}
        & $0.135 \pm 0.072$ \\
        
        15
        & $0.117 \pm 0.042$
        & $0.084 \pm 0.044$
        & $0.109 \pm 0.037$
        & $0.220 \pm 0.085$
        & $0.074 \pm 0.034$
        & $0.148 \pm 0.092$ \\
        
        18
        & $0.086 \pm 0.011$
        & $0.073 \pm 0.024$
        & $0.088 \pm 0.020$
        & $0.253 \pm 0.122$
        & $0.096 \pm 0.042$
        & $0.176 \pm 0.034$ \\

        21
        & $0.062 \pm 0.022$
        & $0.073 \pm 0.051$
        & $0.176 \pm 0.125$
        & $0.142 \pm 0.045$
        & $0.032 \pm 0.008$
        & $0.212 \pm 0.056$ \\
        
        cVAE
        & \textcolor{blue}{$0.244$}
        & \textcolor{blue}{$0.219$}
        & \textcolor{blue}{$0.331$}
        & \textcolor{blue}{$0.537$}
        & $0.164$
        & \textcolor{blue}{$0.379$} \\

        \midrule
        \addlinespace[0.25em]
        
        \textbf{\shortstack{$g(\mathbf{c})$: Embedding\\ dimension} }
        & \multicolumn{6}{c}{
            \textbf{Wasserstein distance (WD) ($\downarrow$)}
        } \\
        \cmidrule(lr){2-7}
        & \textbf{LVEDV}
        & \textbf{RVEDV}
        & \makecell{\textbf{Myocardial}\\\textbf{mass}}
        & \makecell{\textbf{RVEDV/LVEDV}\\\textbf{ratio}}
        & \makecell{\textbf{Long-axis}\\\textbf{length}}
        & \makecell{\textbf{LV}\\\textbf{sphericity}} \\
        \midrule
        
        6
        & $1.042 \pm 0.198$
        & $1.499 \pm 0.112$
        & $0.751 \pm 0.197$
        & $0.940 \pm 0.333$
        & $0.520 \pm 0.153$
        & $0.738 \pm 0.157$ \\
        
        9
        & $1.292 \pm 0.331$
        & $2.082 \pm 0.682$
        & $1.078 \pm 0.205$
        & $1.550 \pm 0.508$
        & \textcolor{blue}{$0.747 \pm 0.283$}
        & $1.010 \pm 0.221$ \\
        
        15
        & $1.409 \pm 0.310$
        & $1.131 \pm 0.104$
        & $1.224 \pm 0.210$
        & $0.912 \pm 0.059$
        & $0.489 \pm 0.038$
        & $0.660 \pm 0.088$ \\
        
        18
        & \textcolor{blue}{$1.431 \pm 0.134$}
        & $1.807 \pm 0.334$
        & $1.037 \pm 0.109$
        & $1.567 \pm 0.027$
        & $0.497 \pm 0.136$
        & \textcolor{blue}{$0.787 \pm 0.090$} \\

        21
        & $1.002 \pm 0.389$
        & \textcolor{blue}{$2.363 \pm 0.254$}
        & $1.362 \pm 0.553$
        & $1.103 \pm 0.400$
        & $0.543 \pm 0.192$
        & $0.568 \pm 0.091$ \\
        
        cVAE
        & $0.899$
        & $2.184$
        & \textcolor{blue}{$1.958$}
        & \textcolor{blue}{$2.734$}
        & $0.425$
        & $0.690$ \\
        
        \bottomrule
    \end{tabular}
    }
\end{table*}

\clearpage
\subsection{Latent dimensionality ablation study}
\label{sec:appendixAblationStudiesLatentDim}
\begin{figure}[h!]
\centering
  \includegraphics[width=\linewidth]{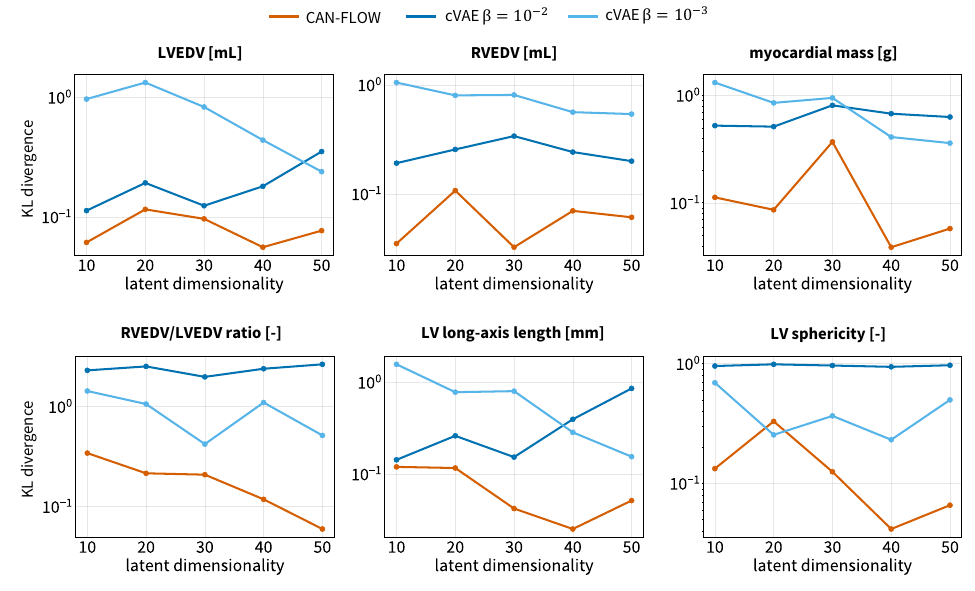}
  \caption{\textbf{Effect of latent space dimensionality on model performance.}
  Performance is quantified using the KL divergence between real and synthetic clinical phenotype distributions. 
  CAN-FLOW is compared with the two best-performing cVAE variants across different latent-space dimensions.
  Lower values indicate better agreement with the real phenotypic distributions.}
  \label{fig:fig_appendix_6}
\end{figure}

We investigate the sensitivity of CAN-FLOW and the two best-performing cVAE variants ($\beta=10^{-2}$, $\beta=10^{-3}$) to the dimensionality of the latent space. 
For each model and latent dimension, we assess how accurately the synthetic cohorts reproduce the real distributions of the derived clinical phenotypes. 
Here, we quantify distributional agreement using the KL divergence, which is closely related to the training objectives of both frameworks. 
Figure~\ref{fig:fig_appendix_6} shows that CAN-FLOW consistently provides a more accurate approximation of the real clinical phenotype distributions than both cVAE variants across all considered latent-space dimensions. 
For the cVAE with $\beta=10^{-3}$, performance often can improve as the latent dimensionality increases. The relatively weaker regularization may allow the model to exploit the additional latent capacity and encode a richer representation of the anatomical variability.
In contrast, the performance of the more strongly regularized cVAE ($\beta=10^{-2}$) can be either relatively insensitive to the latent dimensionality, or can deteriorate as the dimensionality increases. 
A possible explanation is that the stronger regularization constrains the latent posterior more closely to the prior, limiting the model’s ability to use the additional dimensions to encode informative features.
For CAN-FLOW, no clear monotonic relationship between latent dimensionality and performance is observed. Nevertheless, its consistently improved results across the investigated dimensions indicate that the performance advantage of CAN-FLOW is robust to the choice of latent space dimensionality.

\clearpage
\bibliographystyle{elsarticle-num} 
\bibliography{library.bib}
\end{document}